\documentclass[lettersize,journal]{IEEEtran}
\usepackage{amsmath,amsfonts}
\usepackage{algorithmic}
\usepackage{algorithm}
\usepackage{array}
\usepackage[caption=false,font=normalsize,labelfont=sf,textfont=sf]{subfig}
\usepackage{textcomp}
\usepackage{stfloats}
\usepackage{url}
\usepackage{verbatim}
\usepackage{graphicx}
\usepackage{cite}
\usepackage{bm}
\usepackage{booktabs}
\usepackage{multirow}
\usepackage{xcolor}  
\usepackage{soul}
\usepackage{hyperref}
\usepackage{everypage}  

\newcommand{\MyHeader}{%
This article has been accepted for publication in IEEE Transactions on Pattern Analysis and Machine Intelligence. This is the author's version which has not been fully edited and content may change prior to final publication. Citation information: DOI 10.1109/TPAMI.2025.3647835
}

\newcommand{\MyFooter}{%
This work is licensed under a Creative Commons Attribution 4.0 License. For more information, see \url{https://creativecommons.org/licenses/by/4.0/}
}

\AddEverypageHook{%
  \begin{tikzpicture}[remember picture,overlay]
    \node[anchor=north, yshift=-0.5cm] at (current page.north) {\parbox{\textwidth}{\centering\small \MyHeader}};
  \end{tikzpicture}%
  \begin{tikzpicture}[remember picture,overlay]
    \node[anchor=south, yshift=0.5cm] at (current page.south) {\parbox{\textwidth}{\centering\small \MyFooter}};
  \end{tikzpicture}%
}

\usepackage{tikz}

\begin{document}

\title{Iterative Differential Entropy Minimization (IDEM) method for fine rigid pairwise 3D Point Cloud Registration: A Focus on the Metric}

\author{Emmanuele Barberi, Felice Sfravara, and Filippo Cucinotta
\thanks{(Corresponding author: Emmanuele Barberi).
The authors are with Department of Engineering, University of Messina, C.da di Dio Sant'Agata 98166 Messina, Italy. E-mail: (emmanuele.barberi, felice.sfravara, filippo.cucinotta)@unime.it}
\thanks{Manuscript received ...; revised ...}}

\maketitle

\begin{abstract}
Point cloud registration is a central theme in computer vision, with alignment algorithms continuously improving for greater robustness. Commonly used methods evaluate Euclidean distances between point clouds and minimize an objective function, such as Root Mean Square Error (RMSE). However, these approaches are most effective when the point clouds are well-prealigned and issues such as differences in density, noise, holes, and limited overlap can compromise the results. Traditional methods, such as Iterative Closest Point (ICP), require choosing one point cloud as fixed, since Euclidean distances lack commutativity. When only one point cloud has issues, adjustments can be made, but in real scenarios, both point clouds may be affected, often necessitating preprocessing. The authors introduce a novel differential entropy-based metric, designed to serve as the objective function within an optimization framework for fine rigid pairwise 3D point cloud registration, denoted as Iterative Differential Entropy Minimization (IDEM). This metric does not depend on the choice of a fixed point cloud and, during transformations, reveals a clear minimum corresponding to the best alignment. Multiple case studies are conducted, and the results are compared with those obtained using RMSE, Chamfer distance, and Hausdorff distance. The proposed metric proves effective even with density differences, noise, holes, and partial overlap, where RMSE does not always yield optimal alignment. 
\end{abstract}

\begin{IEEEkeywords}
Point cloud, registration, entropy, iterative closest point (ICP), root mean square error (RMSE), noise, density, holes, partial overlap.
\end{IEEEkeywords}

\section{Introduction}
\IEEEPARstart{P}{oint} clouds are a discrete three-dimensional (3D) representation of a given object or scene. In their simplest form, they consist of points uniquely defined by three spatial coordinates (x, y, and z), but they can also contain additional information such as color, intensity, and normals. There are various technologies for acquiring point clouds, including Light Detection and Ranging (LiDAR) systems, structured light 3D scanners, photogrammetry, microscopy, computed tomography (CT), and more. 
Frequently, acquisition technologies require multiple scans from different points of view to obtain a complete representation of a target object or scene. This results in multiple partial point clouds that need to be properly aligned and merged into a common reference system to reproduce the target's geometry. This procedure is known as registration and it has a pivotal role in several application such as 3D reconstruction, mixed reality and Simultaneous Localisation And Mapping (SLAM). Given two point clouds, $P_1$ and $P_2$ which partially overlap, obtained from two different views of the same object, with P   fixed as the reference, the registration process allows for obtaining the transformation matrix that appropriately aligns the target point cloud ($P_2$) with $P_1$. Given the complexity of the required transformation, point clouds registration can be categorized as either rigid or non-rigid and, based on the number of views involved, it can also be classified as pairwise or multi-view registration \cite{1}.
\subsection{Background}
\begin{figure}[h]
\centering
\includegraphics[width=2.3in]{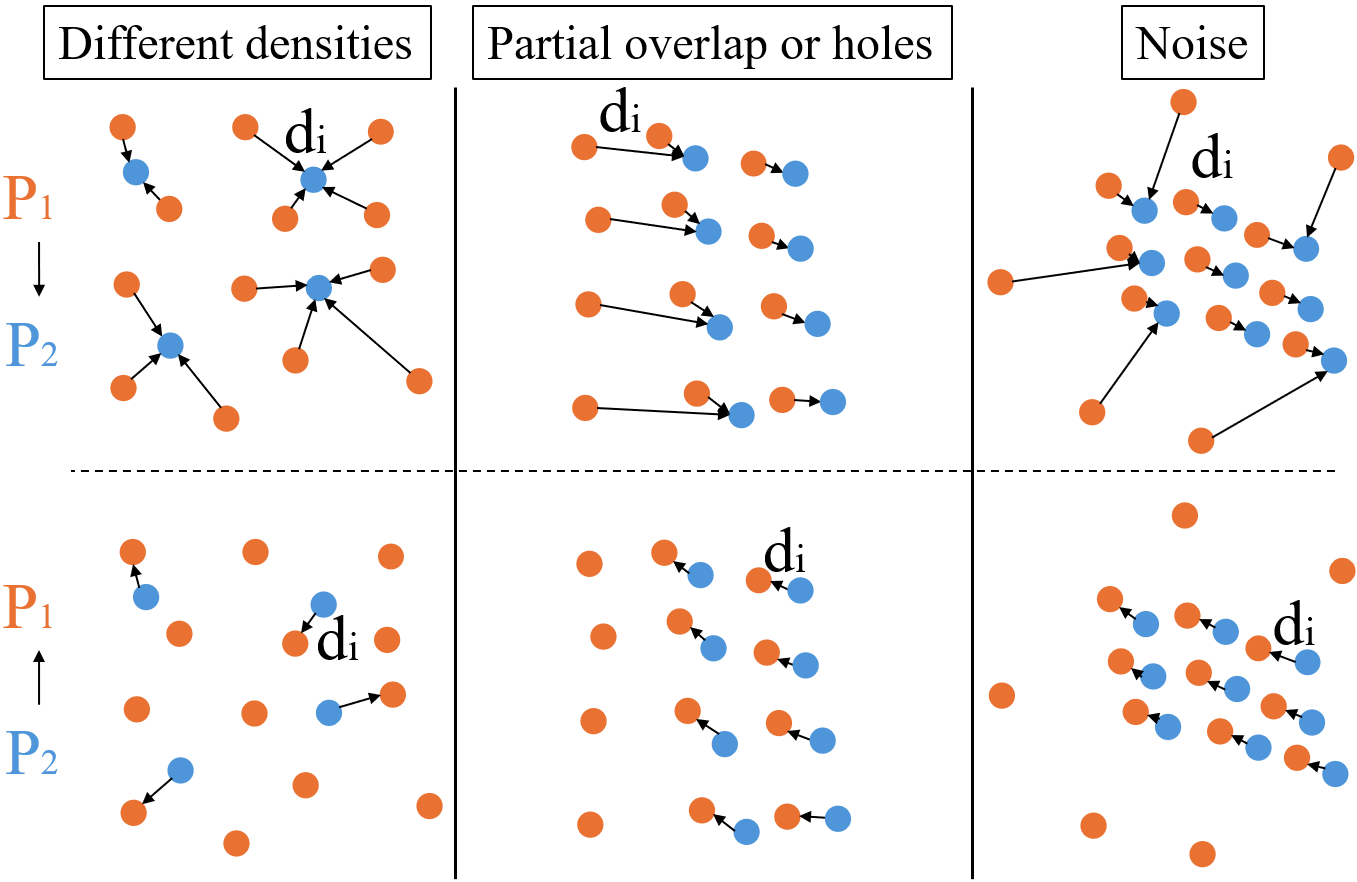}
\caption{Demonstration that the calculation of the Euclidean distance between two point clouds is not commutative.}
\label{F1}
\end{figure}

In scientific literature, there are several registration techniques. Lyu et al. \cite{1} developed ICTaxon, throughout which it is possible to see the existing rigid pairwise point cloud registration methods. These methods are mostly based on Iterative Closest Point (ICP) and its variants \cite{2,3,4,5}, but also on Random SAmple Consensus (RANSAC) \cite{6,7,8,9}, on probabilistic methods \cite{10,11, 12} and more \cite{13,14,15,16}. A significant portion of other registration methods focuses on features extraction and Machine Learning (ML) \cite{17,18,19,20,21,22,23,24}. In particular, the Second Order Spatial Compatibility for Efficient and Robust Point Cloud Registration ($SC^2-PCR$) approach by Chen et al. \cite{25}  introduces a second-order spatial compatibility measure that evaluates correspondence similarity based on global compatibility rather than local consistency. This enables a clearer distinction between inliers and outliers early in the process. The method further employs a global spectral technique to identify reliable correspondences, improving registration accuracy. Qin et al. \cite{26} proposes Geometric Transformer as a robust superpoint matching method that leverages pairwise distances and triplet-wise angles to learn geometric features. Its design ensures invariance to rigid transformations and effectiveness in low-overlap scenarios. The high matching accuracy eliminates the need for RANSAC. Yu et al. \cite{27} introduced RoITr, a Rotation-Invariant Transformer method to address pose variations in point cloud matching, improving stability and accuracy. Unlike traditional deep matchers that rely on data augmentation for rotation invariance, RoITr provides intrinsic rotation invariance. It combines a local attention mechanism with point pair feature coordinates and a global transformer with rotation-invariant cross-frame spatial awareness to enhance feature distinctiveness and robustness in low-overlap scenarios. The method proposed by She et al. \cite{28} enhances point cloud registration robustness to noise and perturbations using graph neural partial differential equations (PDEs) and heat kernel signatures. Graph neural PDEs extract stable features, while heat kernel signatures aid keypoint matching. A learnable Singular Value Decomposition (SVD) module then refines alignment. Experiments show state-of-the-art accuracy with superior resilience to noise and shape distortions.
Other registration algorithms use entropy-based methods, which will be discussed later.
One of the most used and studied point cloud registration methods is the ICP, developed in 1992 by Besl and McKay \cite{2} and Chen and Medioni \cite{3}. It is designed to align or register a wide variety of objects, including point clouds, poly-lines and surfaces. 
ICP is an iterative method through which, iteration after iteration, it is possible to align two point clouds. To achieve the alignment, at each iteration, the distance between the point clouds is minimized using an objective function, which is typically based on the least squares method and its variants.
More specifically, given two point clouds $P_1$ and $P_2$, where $p_{1_i}$ are the points in $P_1$ ($p_{1_i} \in P_1$) and $p_{2_j}$ are the points in $P_2$ ($p_{2_j} \in P_2$), one of the point clouds is initially chosen as the model (or reference), for example $P_1$, while the other point cloud is designated as the target ($P_2$). These are also referred to as fixed and moving, respectively. The algorithm then proceeds as follows:
\begin{enumerate}
\item{A point-to-point correspondence is established, meaning for each point in $P_2$, the corresponding point in $P_1$ is found. This correspondence is determined by calculating the Euclidean distance: for each $p_{2_j} \in P_2$, the nearest $p_{1_i} \in P_1$ is selected.}
\item{Rigid transformations are iteratively applied to $P_2$ in order to minimize the chosen objective function. The iterations can be stopped after a predefined number of steps or when the objective function reaches a value deemed satisfactory.}
\end{enumerate}
The output is a transformation matrix (\textbf{T}), containing the elements of the rotation matrix \textbf{R} ($r_{i,j}$) and those of the translation vector \(\bm{t}\) (\textbf{$t_i$}), that allows $P_2$ to undergo a rigid roto-translation from its initial position, aligning it with $P_1$. In general, through \textbf{T}, a point in space 
 \ensuremath{\bm{x}}$(x_1,x_2,x_3)$ can be transformed into the point \ensuremath{\bm{x}'}$(x'_1,x'_2,x'_3)$ by means of (\ref{Eq1}).
\newcommand{\vect}[1]{\bm{#1}} 
\begin{equation}
\label{Eq1}
\ensuremath{\bm{x}'} = \mathbf{T} \ensuremath{\bm{x}} = \mathbf{R} \ensuremath{\bm{x}} + \vect{t}.
\end{equation}
It is also important to highlight that the iterative process is influenced by the method used to find the correspondences between $p_{1_i}$ and $p_{2_j}$ \cite{29}. In fact, there are primarily three methods: point-to-point \cite{30}, point-to-line \cite{31,32}, and point-to-plane \cite{33}.
Over the years, ICP has been extensively applied and studied, leading to the development of numerous variants to enhance its performance. For instance, the study of Hong et al. \cite{34} presents an improvement of the traditional ICP algorithm, called Velocity updating Iterative Closest Point (VICP). It integrates velocity updates and compensating scans simultaneously to enhance the convergence of the alignment between sets of points, making it particularly useful in robotics. Zhang et al. \cite{35} propose a faster and more robust ICP optimized for handling large 3D data sets. The approach is based on Welsch’s function and Amderson acceleration. The work of Du et al. \cite{36} introduces the probability ICP method, designed to handle multidimensional point sets affected by noise, thereby improving registration accuracy. 
The need to improve the performance of ICP arises from the fact that this method, among others, has certain limitations. Indeed, scientific literature \cite{35,36,37,38,39,40,41} indicates that the quality of point cloud registration is affected by factors such as differences in density between point clouds. An additional factor is the non-commutativity of Euclidean distance calculation between two point clouds \cite{42}. Since registration methods based on Euclidean distance calculation require choosing one point cloud as the target and the other as the source, the non-commutative nature of this distance calculation affects the result of the registration.    
As a demonstration of the last point, Fig.~\ref{F1} illustrates the non-commutativity of distance calculations between point clouds with varying densities, partial overlap point clouds (or presence of holes), and presence of noise. Let $P_1$ and $P_2$ represent two point clouds (for simplicity, the two-dimensional (2D) case is shown). 
The distances are computed as follows: in the first case (top), for each point in $P_1$, the distance to the nearest point in $P_2$ is calculated, with the sense of the distance calculation being $P_1$→$P_2$. In the second case (bottom), the reverse is applied, with the sense of the calculation now being $P_2$→$P_1$. The distances $d_i$ and $d_j$ are different from each other. The process of calculating distances (i.e., finding the nearest point in the other point cloud) does not follow commutative properties \cite{42}.   For these reasons, metrics based on the calculation of distances are influenced by the order in which the point clouds are chosen, an improper selection of the reference point cloud can result in Euclidean distance evaluations between corresponding points that are often inaccurate.
To mitigate these effects, certain precautions are commonly adopted, which are also used in conducting geometric deviation analyses. This technique is employed to evaluate differences, for instance, between a product and its original computational model in order to assess geometric discrepancies, i.e., the fidelity of the realization. The precautions to be taken involve the selection of the model point cloud and the target point cloud. Specifically:
\begin{itemize}
\item{When dealing with point clouds of different densities, distances should be computed from the less dense point cloud to the denser one;}
\item{In cases of partially overlapping point clouds, distances should be computed from the smaller point cloud to the larger one;}
\item{In the presence of noise in one point cloud, distances should be computed from the noise-free point cloud.}
\end{itemize}
However, considering that in real-world scenarios point clouds may have varying local densities and both may contain holes and noise, selecting the reference point cloud is not always straightforward. 
The reason why the quality of point cloud registration using ICP can be compromised by the aforementioned issues lies in the sensitivity of ICP to local minima \cite{43}.
Consider the ideal case of two identical point clouds being perfectly aligned; in this scenario, they would overlap perfectly, point by point. A commonly used evaluation metric, such as Root Mean Square Error (RMSE) (\ref{Eq2}), would confirm this alignment. 
\begin{equation}
\text{RMSE} = \sqrt{\frac{1}{n} \sum_{i=1}^{n} \left| \mathbf{R} \vect{p_{2_j}} + \vect{t} - \vect{p_{1_i}} \right|^2 }.
\label{Eq2}
\end{equation}
In the RMSE’s formulation, which is widely employed to assess the performance of ICP \cite{44}, the term $n$ indicates the number of the $p_{1_i} \in P_1$. In this specific case, where identical point clouds are in complete overlap, the RMSE value is zero. This represents the global minimum and is consistent with the expected outcome (perfect alignment). However, methods like ICP, which rely on performance metrics such as this one, based on Euclidean distance, are sensitive to local minima \cite{43}. This means that the algorithm might find a solution it considers optimal, but which does not correspond to the desired result, as it may converge to a local minimum in the neighborhood of the transformations performed, rather than the actual global minimum.
This underscores the need for preprocessing the data before applying ICP to a point cloud registration problem. What is required is to bring the target point cloud as close as possible to the reference point cloud through an initial transformation. Therefore, a pre-alignment of the point clouds is necessary to achieve a robust result \cite{45}. This pre-alignment can be performed either manually or automatically. Thus, the accuracy of the point cloud registration using ICP heavily depends on the initial alignment conditions of the point clouds.  
\subsection{Entropy-based point cloud registration}
Other works on point cloud registration adopt entropy-based methods. Zhan et al. \cite{46} developed the Entropy and Particle Swarm Algorithm (EPSA). In their approach, noise present in point clouds is removed by finding the nearest points using a k-d tree, while entropy is evaluated by considering the probability of occurrence of the projected points along each spatial axis in relation to the frequency distribution. Chen et al. \cite{47} introduces a method combining a terrestrial laser scanner and a smartphone for the coarse registration of leveled point clouds with small roll, pitch angles, and height differences. The approach starts by calculating the distance between two scan positions using GPS coordinates. Then, 2D distribution entropy measures the coherence between scans to find initial transformation parameters. The Iterative Minimum Entropy (IME) method is proposed to refine these parameters, using two criteria: the difference between average and minimum entropy, and the deviation from the expected entropy.

Entropy-based methods are often associated with ICP   in the initial phase to remove points that could compromise its robustness. In this context, the study by Donoso et al. \cite{48} introduces three methods to improve ICP for terrain registration. Two entropy-based methods, eigentropy filtering and unilateral eigentropy rejection, enhance point selection and pair rejection by addressing geometric disorder. The third method, point matching by normal deviation, uses surface normals for more accurate point matching. The experiments demonstrate significant improvements in accuracy, precision, and efficiency, highlighting the effectiveness of eigentropy in making ICP more robust and context-independent across various terrain scenarios.
\subsection{Aim of the Work}
Given the limitations of ICP discussed above, particularly those related to the use of Euclidean distance, this paper aims to propose a new metric based on differential entropy, which can be employed as the core objective function in an optimization framework for fine rigid pairwise point cloud registration, referred to as Iterative Differential Entropy Minimization (IDEM).

The main contributions of this paper can be summarized as follows:
\begin{itemize}
    \item A new metric for rigid pairwise point cloud registration based on differential entropy is proposed.
    \item The proposed metric overcomes several limitations of ICP arising from the use of Euclidean distances, including robustness to holes, noise, varying point cloud densities, and partial overlap. It is independent of the choice of the reference point cloud, exhibiting a commutative behavior, and demonstrates a clearly marked minimum corresponding to perfect alignment. It is also invariant to rigid rotations and translations of point clouds.
    \item A comparative analysis with RMSE, Hausdorff distance, and Chamfer distance is carried out, highlighting the advantages of the proposed metric.
    \item A sensitivity analysis based on a Monte Carlo approach is performed to assess the robustness of the proposed metric.
\end{itemize}

\subsection{Outline}
After an introductory section clarifying the context and providing a bibliographic analysis of the state of the art, Section \ref{sec:2} introduces the method, outlining the theoretical aspects underlying the proposed approach. This is followed by the definition of the proposed metric and all the necessary mathematical details for the implementation of the method. Section \ref{sec:3} is dedicated to the conducted experiments, starting with a list of the experiments carried out. This section concludes with the sensitivity analysis performed on the proposed metric, followed by a summary of the experiments conducted and the results obtained. Finally, Section \ref{sec:4} presents the conclusions.

\section{Materials and Methods}\label{sec:2}
\subsection{Introduction to the Method}
Differential entropy is a concept introduced by Claude Shannon in 1948 \cite{49}, and for this reason, it is often referred to as Shannon entropy. It quantifies the information contained in a given event. Shannon entropy, expressed as $H(X)$, represents the amount of information contained in the discrete random variable $X$, which has its own probability distribution. Depending on the type of distribution, differential entropy takes on different formulations. Specifically, in this work, reference is made to a multivariate normal distribution, and the corresponding formulation of Shannon entropy (expressed in nats) is as follows:
\begin{equation}
\mathit{Differential} \mathit{Entropy} = \frac{1}{2} \ln \left( (2 \pi e)^N \,\det(\Sigma) \right) ,
\label{Eq3}
\end{equation}
given $N$ as the dimensionality of the dataset and $\Sigma$ as the covariance matrix of the data. Its determinant, often referred to as \textit{“generalised variance”}, it was considered by Wilks \cite{50} as \textit{“a scalar measure of overall multidimensional dispersion”} \cite{51}.
In the context of point clouds, calculating differential entropy provides insights into the distribution of points. Naturally, the entropy value must be interpreted within the specific context under study, but in general, the more concentrated the points are in a given area, the lower the differential entropy. Conversely, as the points become more spread out, the entropy value tends to increase. The determinant of the covariance matrix plays a crucial role. It is, in fact, an invariant, meaning its value will not be affected by the rigid rotation and translation between the two point clouds. Additionally, it is given by the product of the eigenvalues of the covariance matrix, and thus provides quantitative information on how the points are distributed in space. Given this, it is reasonable to infer that similar point clouds, representing the same object, should exhibit similar values of differential entropy. This principle underpins the work of Adolfsson et al. \cite{52}, who introduced a method (CorAl) for evaluating the alignment of point clouds acquired via LiDAR. The alignment of two point clouds is evaluated by introducing a numerical quality index, which approaches zero in the case of perfect alignment. The differential entropy of each point cloud is calculated using (\ref{Eq3}). Specifically, for each point in the cloud, its entropy is determined along with the points within a defined spherical neighborhood, by computing the covariance matrix of those points. The sum of the differential entropy values for all points provides the overall entropy of the point cloud. The quality of the alignment is then obtained by subtracting the sum of the individual point clouds' entropies from the entropy of the combined (joint) point cloud. Additionally, it is possible to plot the specific quality index value for each point using a colormap of the joint point cloud, allowing areas with greater deviations to be identified.
Previous studies relied on CorAl to develop the Differential Entropy for Deviation Analysis (\textit{DEDA}) method \cite{53,54}. In these studies a robustness analysis of the method highlighted its advantages over the classical approach based on Euclidean distance. The robustness analysis conducted in these studies involved using two identical point clouds and evaluating the deviation across six   cases: perfectly overlapping point clouds (the reference case), presence of noise or background, presence of holes, different densities of the point clouds, and actual misalignment (translation and rotation).
In the reference case, as expected, a quality index of 0 was obtained, indicating perfect overlap. The same optimal value is achieved in in presence of noise or background. This occurs due to the way the problem is formulated: the background points belong to the modified point cloud, and locally, the values of the entropies, as defined, coincide, making their difference (i.e., the entropy value of the noise) equal to zero. The entropic contribution of the noise can also be nullified by introducing 'if' conditions, setting a lower limit on the number of points that must be contained within $\rho$.
In the presence of holes, the entropic contribution of the non-overlapping regions is zero, except at the interface, generating only a boundary effect whose intensity is relatively low and depends on the search radius $r$. Conversely, the classical deviation analysis (based on calculating mean and standard deviation) shows non-zero values. The quality index is not zero but two or three orders of magnitude lower than those in the case of misalignment. As regard the point clouds with differences in densities, the quality index is also non-zero but is one or two orders of magnitude lower than in cases of actual misalignment.
The results from these previous studies highlight the method's immunity to noise, its slight sensitivity to holes, and its moderate sensitivity to differing point cloud densities. Furthermore, another key strength of this method is its independence from having to choose a reference or target point cloud, unlike classical methods based on Euclidean distance. Building on the insights presented thus far, a differential entropy-based metric is proposed, which can be implemented in an iterative method for fine rigid pairwise point cloud registration.

\subsection{IDEM – Iterative Differential Entropy Minimization}
Modifications were made to the formulation of differential entropy for a multivariate Gaussian distribution (\ref{Eq4}), which can also be found in another work \cite{55} by some of the authors of this paper, where the Differential Entropy-based Compactness Index (\textit{DECI}) was introduced. The new formulation is as follows:
\begin{equation}
\mathit{h_i(p_k)} = \frac{1}{2} \ln \left( (2 \pi e)^N \,\det(\Sigma (p_k)\right) + 1 ) ,
\label{Eq4}
\end{equation}
where $h_i$ represents the differential entropy of the $i$-th point in a given point cloud ($P$), and $p_k$ is a subset of $P$ obtained by identifying a spherical neighborhood $\rho$ with a radius $r$. Here, $k$ refers to the number of points within this neighborhood (Fig.~\ref{F3}).
\begin{figure}[h!]
\centering
\includegraphics[width=1.2in]{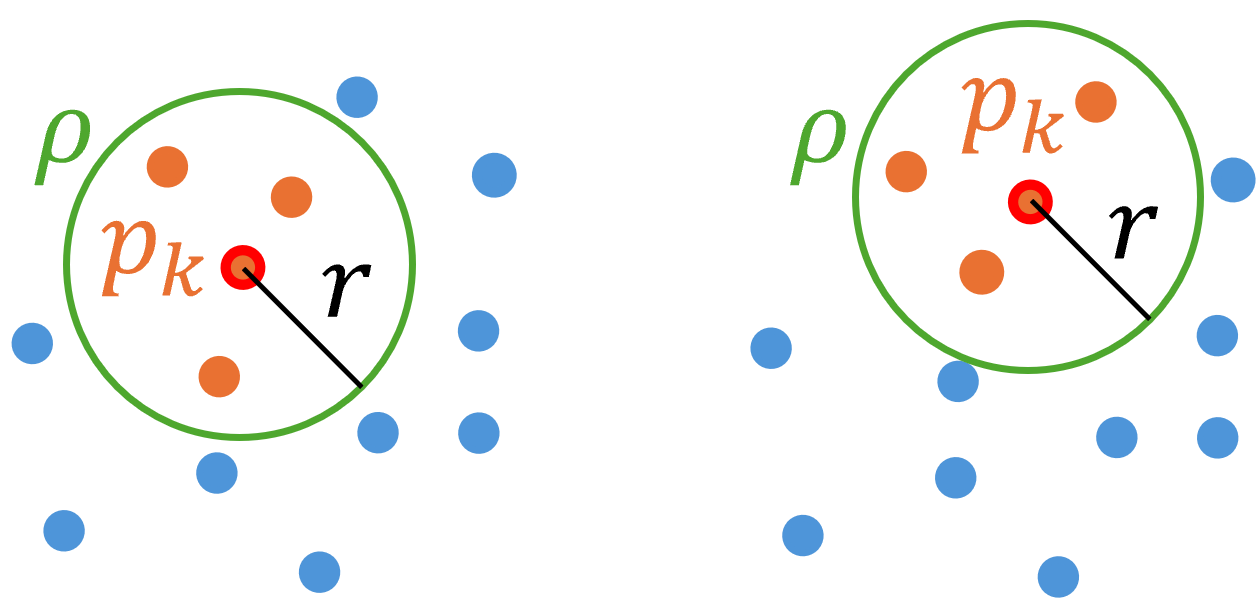}
\caption{2D representation of the identification of the neighborhood $\rho$, radius $r$, and the set $p_k$.}
\label{F2}
\end{figure}
The addition of the term $+1$ presents several advantages. For instance, in the 3D case, when the number of points within the neighborhood $\rho$ is less than or equal to 3 ($k \leq 3$), the determinant of the covariance matrix ($\Sigma$) becomes zero. This results in the logarithm of 1 in the formulation, which, being equal to 0, nullifies the entropic contribution of the $i$-th point. Furthermore, in cases where the determinant is non-zero, the argument of the logarithm will always be greater than 1, ensuring that differential entropy values remain positive.
Thus, the overall differential entropy of a point cloud $P$ ($H(P)$) will be:
\begin{equation}
\label{Eq5}
\mathit{H(P)} = \sum_{i=1}^{n_p} h_i(p_k),
\end{equation}
where $n_P$ is the number of the point of $P$.

The proposed method is presented below.
Let $P_1$ and $P_2$   be two point clouds to be registered into a joint point cloud $P_J$, which is formed by points $\bm{P_1} \in P_1$ and $\bm{P_2} \in P_2$, such that
\begin{equation}
\bm{P_J} = \begin{bmatrix} \bm{P_1} \\ \bm{P_2} \end{bmatrix} \quad
\label{Eq6}
\end{equation}
that is, the matrix $\bm{P_J}$ contains, in sequence, the matrices $\bm{P_1}$ and $\bm{P_2}$. 
Let $\bm{h_1}$, $\bm{h_2}$, and $\bm{h_J}$ denote column vectors containing of differential entropy values (as defined in (\ref{Eq4})) of $\bm{P_1}$, $\bm{P_2}$, and $\bm{P_J}$, respectively. Then:
\begin{equation}
\bm{q}(p_k) = \bm{h_J}(p_k) - \begin{bmatrix} \bm{h_1}(p_k) \\ \bm{h_2}(p_k) \end{bmatrix} .
\label{Eq7}
\end{equation}
Based on (\ref{Eq7}), the metric proposed in this paper for the fine rigid pairwise point cloud registration is:
\begin{equation}
q_{tot} = \sum_{i=1}^{|q|} q_i \left( p_k \right) ,
\label{Eq8}
\end{equation} 
given that $|q|$ is the number of elements of \bm{$q(p_k)$} and $q_i (p_k)$ represents its elements.  
As defined, $q_{tot}$ can reach two absolute minima when considering two perfectly overlapping point clouds. Specifically, $q_{tot}$ takes a value of 0 (indicating the minimum) either when the point clouds are perfectly superimposed or when they are separate. In the latter case, each neighborhood $p_k$, evaluated for every point, includes only points from a single point cloud.  
Since the evaluation of $q_{tot}$ depends primarily on the size of $\rho$, specific studies on the search radius $(r)$ size were considered necessary to determine an optimal value. This investigation can be regarded as an ablation study aimed at analyzing the influence of the neighborhood radius on the overall behavior of the proposed metric. Changing the radius affects the number of points that can potentially fall within the neighborhood $\rho$.  

To determine an optimal neighborhood size, the following considerations were made. The minimum number of points within the neighborhood should be greater than 3, as three points in space always lie on a plane and would make the entropy value zero, according to (\ref{Eq4}), since the determinant of the covariance matrix would be zero, thus losing the information of that set of points. The radius also cannot be too large, for example, by encompassing the entire point cloud, for several reasons. Indeed, if the radius were infinitely large, all points would have the same entropy value. This would make the calculation useless and computationally more expensive. Additionally, as the search radius increases, the entropy value will be more influenced by the extent of the cloud than the misalignment observed. In fact, in the case of fine registration, misalignments are presumed to be very small compared to the extent of the point cloud. Therefore, it was concluded that a radius containing at least four points is the smallest one to avoid information loss and should have an extent comparable to the misalignments between the two point clouds. To verify this hypothesis, several analyses were conducted. Initially, the average distance between points in each point cloud was assessed.

Thus, the average distance between points in each point cloud was assessed. More specifically, considering the previously mentioned determinant of the covariance matrix (\ref{Eq9}), the point clouds studied were analyzed by calculating the average value of the distances ($r_{\text{4th}}$) from each point to its fourth nearest neighbor ($r_{4th}$). This is because the fourth nearest neighbor is the one that contributes a non-zero entropy value. It should be noted that the calculation includes the three points closest to the central point. It is reasonable to assume that a radius lower than $r_{\text{4th}}$ would result in a loss of information. Finally, a dimensionless parameter $a$ was introduced:
\begin{equation}
a = \frac{r}{r_{4th}} .
\label{Eq9} 
\end{equation}
The behavior of $q_{tot}$ during a translation along the X-axis can be plotted as a function of $a$, which is a parameter defined by the ratio between the search radius used for calculating differential entropy and the distance $r_{4th}$. 
In cases where the point clouds differ in density and/or number of points, the $r_{4th}$ value will vary for each point cloud. Therefore, in the various experiments presented, the distance $r_{4th}$ was calculated by considering the weighted average between the $r_{4th}$ of the two point clouds to be aligned, where the weights are the percentage of points relative to the total points in the two point clouds. More specifically, the $r_{4th}$ of one point cloud was multiplied by the percentage of points in the other point cloud and vice versa. In this way, in the presence of a less dense point cloud, it has a greater weight in determining the value of the $r$, thereby avoiding information loss.

The point cloud examined in this paper is the Stanford Bunny \cite{57}, hereafter referred to as $B_0$ (Fig.~\ref{F3}) consisting of 1,597 points.

\begin{figure}[h!]
\centering
\includegraphics[width=1.3in]{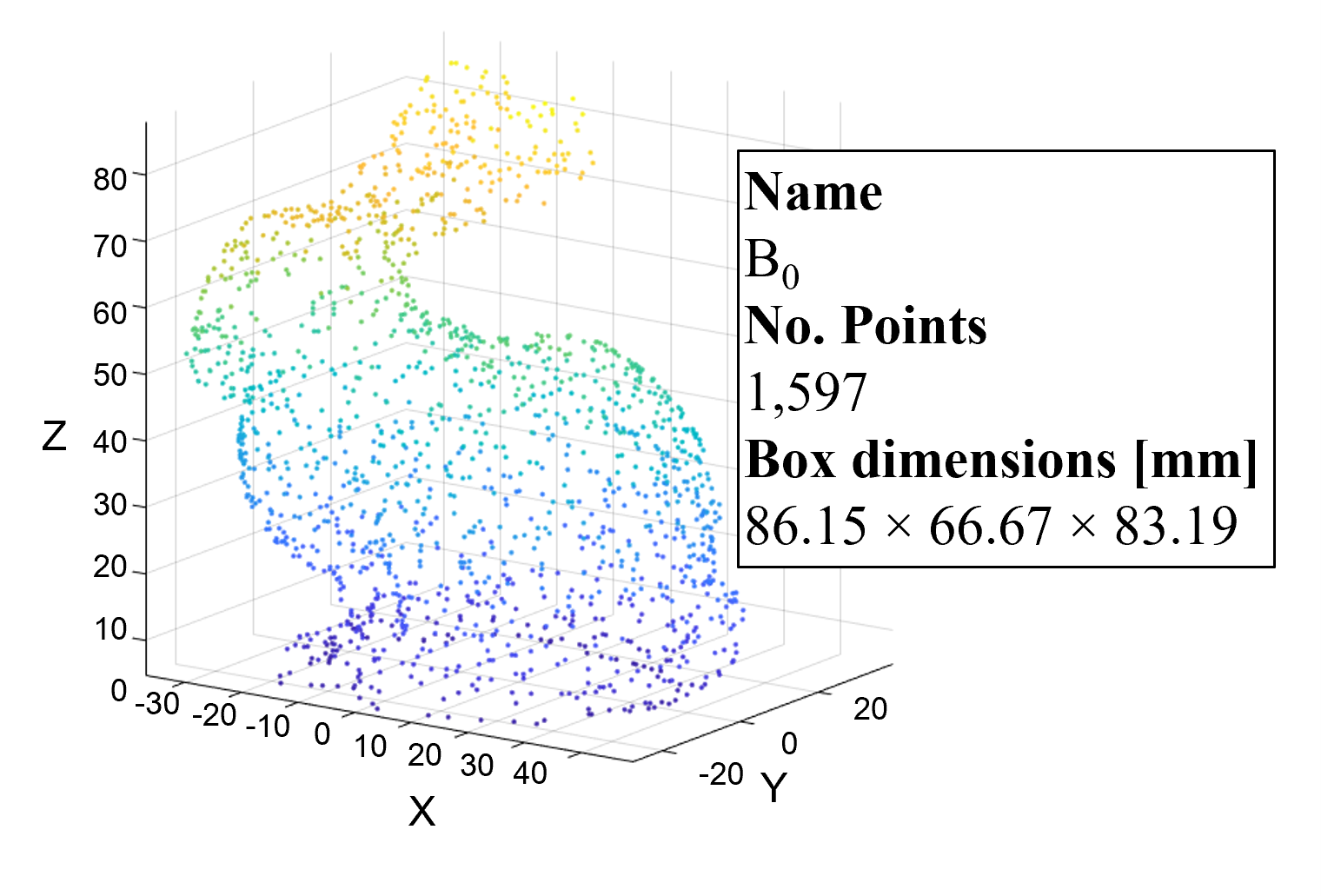}
\caption{Point cloud ($B_0$) used as an example for the experiments.}
\label{F3}
\end{figure}
\begin{figure}[h!]
\centering
\includegraphics[width=\columnwidth]{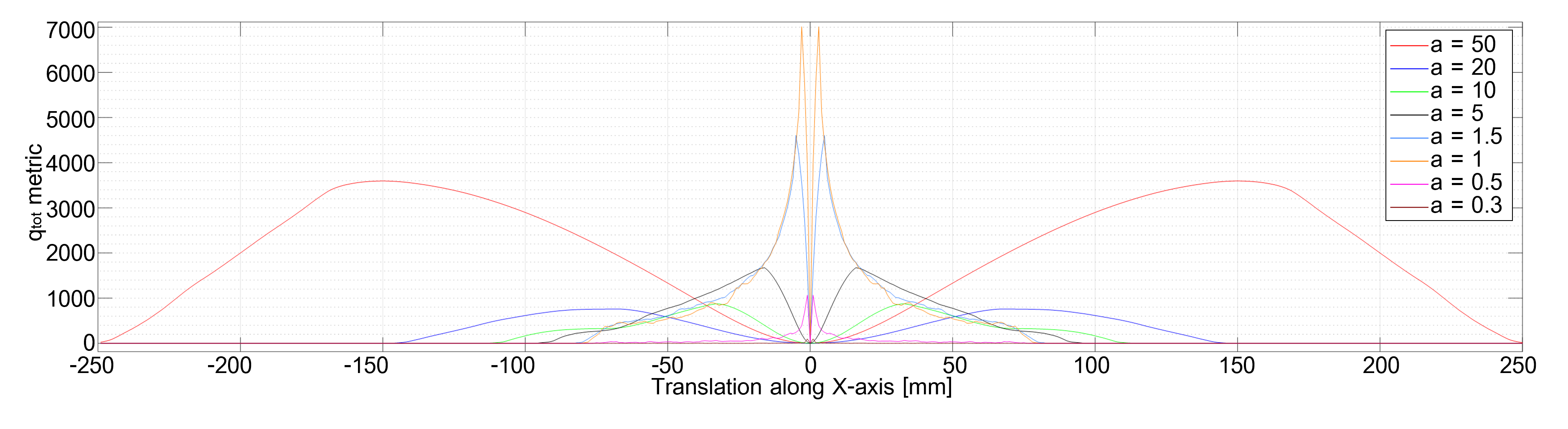}
\caption{Behavior of $q_{tot}$ (translation along the X-axis) as the parameter $a$ varies.}
\label{F4}
\end{figure}

Fig. \ref{F4} illustrates the value of $q_{tot}$ during the translation along the X-axis with a step of 1 mm, varying the parameter $a$, simulating the alignment process of two identical point clouds.
The point 0 on the graph represents the perfect overlap of the two point clouds, and as $a$ varies, $q_{tot}$ is zero, just as it is when there is no overlap between the point clouds. What emerges is that the curve with a value of $a=1$ possesses the highest $q_{tot}$. As $a$ increases (i.e., to values of $r$ which encompasses the entire point cloud to which the center of the neighborhood belongs), the maximum value of $q_{tot}$ initially decreases and then tends to rise again, while the distance between the two peaks, calculated along the X-axis, increases, making the drop towards the minimum ($x=0$) less abrupt. In this case, the behavior of $q_{tot}$ resembles that of a double Gaussian distribution. This behavior has also been observed in other studies based on entropy found in the scientific literature \cite{47}. 

For values of $a$ less than 1, it is noted that the two peaks remain, but their intensity gradually diminishes. This is because reducing $r$ results in fewer points falling within each identified neighborhood, eventually dropping below the threshold of four points. Moreover, the entropic contribution becomes null, leading to a decreasing number of non-zero elements in \bm{$q(p_k)$}, and as $r$ decreases, the summation tends towards 0. Moreover, the decreasing trend exhibited by the curve outside of its two peaks is since, during relative translation, the overlap between the point clouds gradually decreases. Consequently, the points with non-zero entropy values that effectively contribute to the total $q_{tot}$ value become progressively fewer. 

\begin{figure}[hb!]
\centering
\includegraphics[width=\columnwidth]{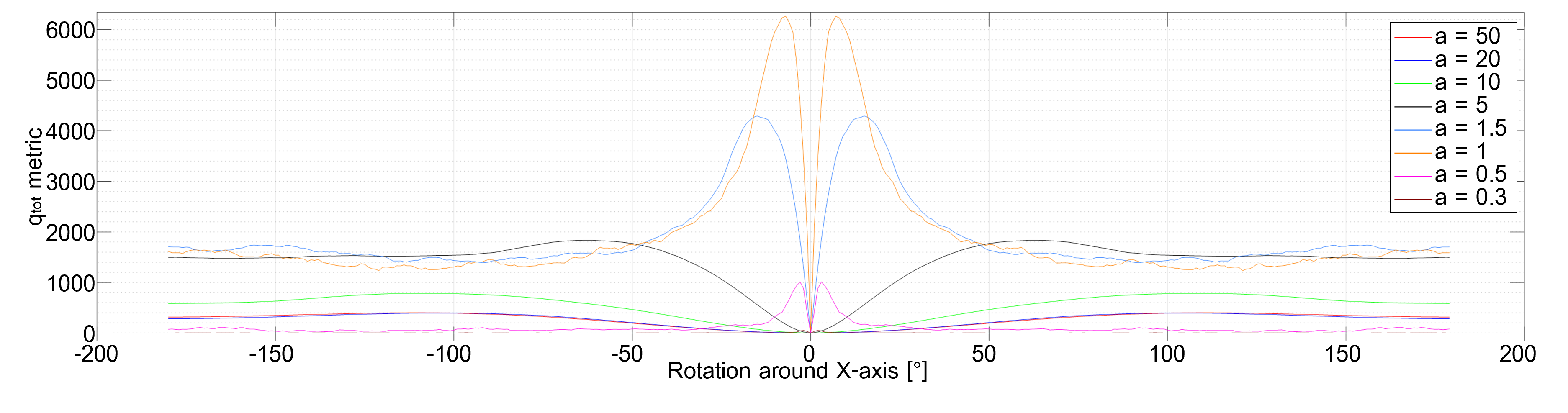}
\caption{Behavior of $q_{tot}$ (rotation around the centroids Z-axis) as the parameter $a$ varies.}
\label{F5}
\end{figure}

Below (Fig.~\ref{F5}) is the trend of $q_{tot}$ for $B_0$, as the rotation around the centroids Z-axis varies.

\begin{figure}[h!] 
    \centering
    \includegraphics[width=\columnwidth]{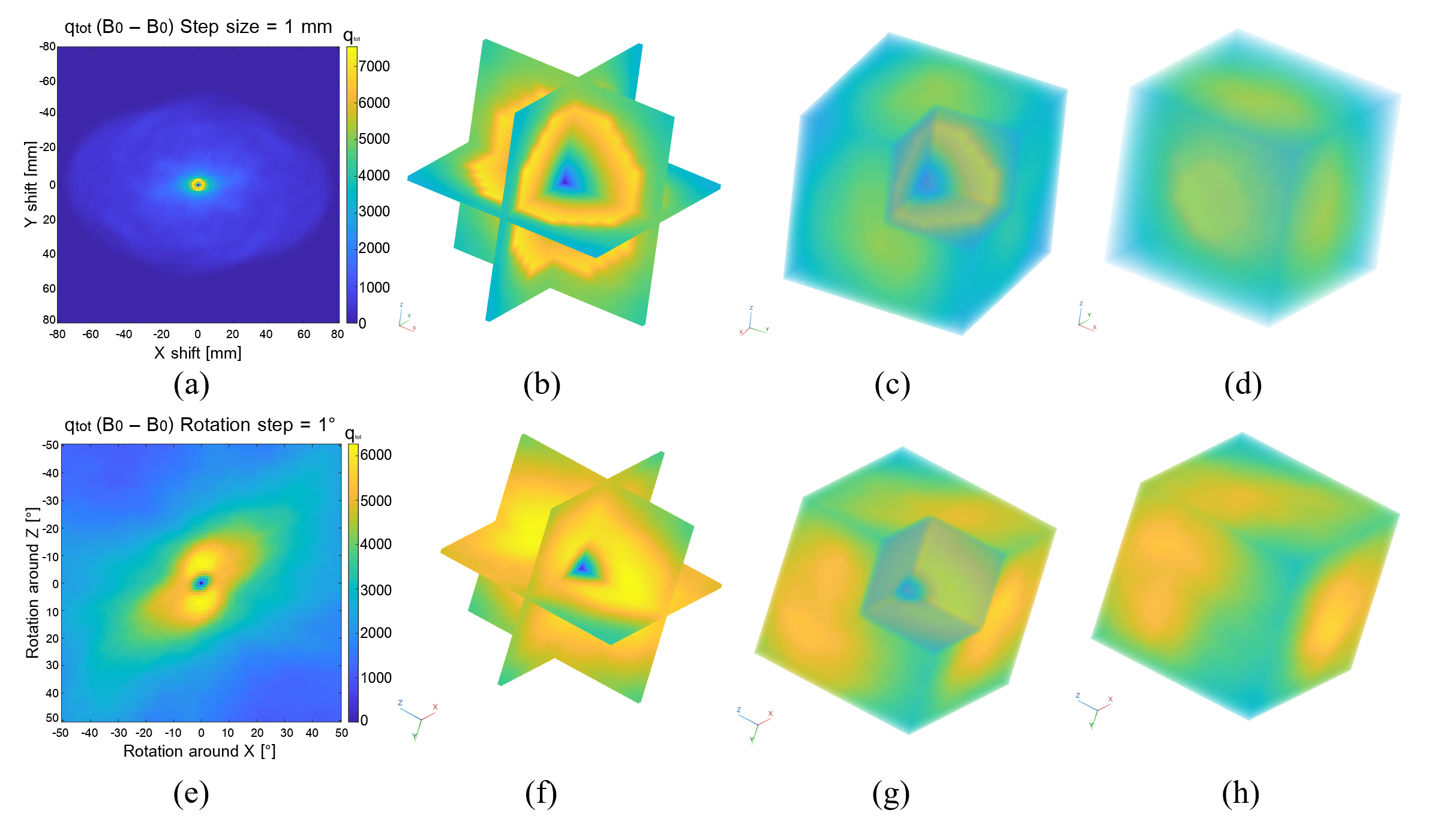} 
    \caption{Colormap of $q_{tot}$ as a function of translations: in the XY plane (a), displacements within the ROI on the XY, XZ, and YZ planes (b), cross-section of the ROI volume (c), and the full volume of the ROI (d). Colormap of $q_{tot}$ as a function of rotations: around the centroids Z-axis (e), rotations within the ROI around the centroids X, Y, and Z axes (f), and the full volume of the ROI (g).}
    \label{F6}
\end{figure}

The position at 0° represents the two point clouds perfectly overlapping. The rotation applied ranges from -180° to +180°. In the displayed graph, the minimum value of $q_{tot}$ represents perfect alignment. Again, the curve with $a=1$ corresponds to the highest $q_{tot}$.

The test conducted with roto-translations shows that the most effective radius is always obtained with a parameter of $a=1$, as it corresponds to the highest peak within a sufficiently narrow range of perfect alignment, and furthermore, adopting this value does not lead to a significant loss of information. The dimensionlessness of the parameter $a$ also ensures that the calculation remains unaffected by changes in the size of the point cloud.

In summary, the study on $r$ reveals that across all values of $a$, the point with the lowest $q_{tot}$ (excluding the trivial case where the clouds are entirely separated) occurs at $x=0$ or $x=0^{\circ}$, corresponding to perfect alignment. Furthermore, as $a$ approaches 1, the $q_{tot}$ peak reaches its highest value. Therefore, choosing a value of $r$ equal to $r_{4th}$ would simplify the configuration of a minimum-search algorithm within the funnel formed by the peaks. 
It has been noted that the best alignment represents a local minimum of $q_{tot}$ and it is always within the peaks.
The area between these two identified peaks thus represents a Region of Interest (ROI) where the experiments will primarily focus.

Therefore, the IDEM method proposed by the authors in this paper involves minimizing $q_{tot}$.
\begin{equation}
q_{IDEM} = \min  (q_{tot})_{ROI} ,
\label{Eq14} 
\end{equation}
where $q_{IDEM}$ is defined as the minimum value of $q_{tot}$ evaluated within the ROI, located between the peaks of the function.

To demonstrate the effectiveness of the method, and to observe what happens on a broader scale, it is also possible to plot 2D maps displaying the values of $q_{tot}$ following translations and rotations between two point clouds. Similarly, RMSE maps can be plotted. The colormaps are set up such that the central pixel (coordinates 0,0) represents perfect alignment. 

In Fig.~\ref{F6}, an example of $q_{tot}$ maps for two identical point clouds ($B_0$) subjected to translation in the XY plane (Fig.~\ref{F6}a) and rotation around the X and Y axes (Fig.~\ref{F6}e) is shown. For clearer visualization, 3D colormaps were also generated, to better highlight the central section of the previous images. They illustrate the behavior of $q_{tot}$ as translations and relative rotations between the two point clouds vary, respectively, in the XYZ (Fig.~\ref{F6}b, Fig.~\ref{F6}c, and Fig.~\ref{F6}d) space and around the X, Y, and Z axes (Fig.~\ref{F6}f, Fig.~\ref{F6}g, and Fig.~\ref{F6}h). In particular, Fig.~\ref{F6}b and Fig.~\ref{F6}f show the $q_{tot}$ across the three planes, Fig.~\ref{F6}c and Fig.~\ref{F6}g provide a sectional rendering of the ROI, while Fig.~\ref{F6}d and Fig.~\ref{F6}h display the complete 3D ROI.

In both cases, the central pixel (perfect alignment) corresponds to a $q_{tot}$ value of zero. It can be observed that, in the case of translation (Fig.~\ref{F6}a), $q_{tot}$ also assumes a zero value when the translating point cloud fully separates from the other point cloud (in accordance with the definitions of differential entropy). Conversely, in the case of rotations (Fig.~\ref{F6}e), the absolute minimum occurs exclusively at the central pixel. 
\section{Experiments}\label{sec:3}
The experimental section of the paper aims to highlight the characteristics of the proposed metric and to showcase its strengths and weaknesses by comparing it with traditional alignment evaluation metrics based on Euclidean distance (\ref{Eq2}). The robustness of the proposed metric for registration will be evaluated and compared with RMSE through various case studies:
\begin{enumerate}
    \item Point clouds with different densities
    \item Presence of noise
    \item Presence of holes
    \item Identical point clouds that are partially overlapping
    \item Similar point clouds
    \item Similar point clouds that are partially overlapping
\end{enumerate}
The experiments were conducted starting from two point clouds in a perfectly aligned position, translating and rotating one of the point clouds. The values of $q_{tot}$ and RMSE were recorded to plot graphs and colormaps that illustrate the point of best alignment according to the studied metrics ($q_{tot}$), compared with the ground truth. It should be noted from the outset that the method proposed in this article is not influenced by the choice of the point cloud to be transformed. The method is entirely independent of the selection of the model point cloud and the target point cloud and, therefore, can be said to possess the commutative property, unlike methods based on Euclidean distance calculations. The entire computational section, where not otherwise specified, including the generation of images and graphs, was implemented using MATLAB \cite{56}.
\subsection{Influence of Point Cloud Density on $q_{tot}$}
\begin{figure}[b] 
    \centering
    \includegraphics[width=\columnwidth]{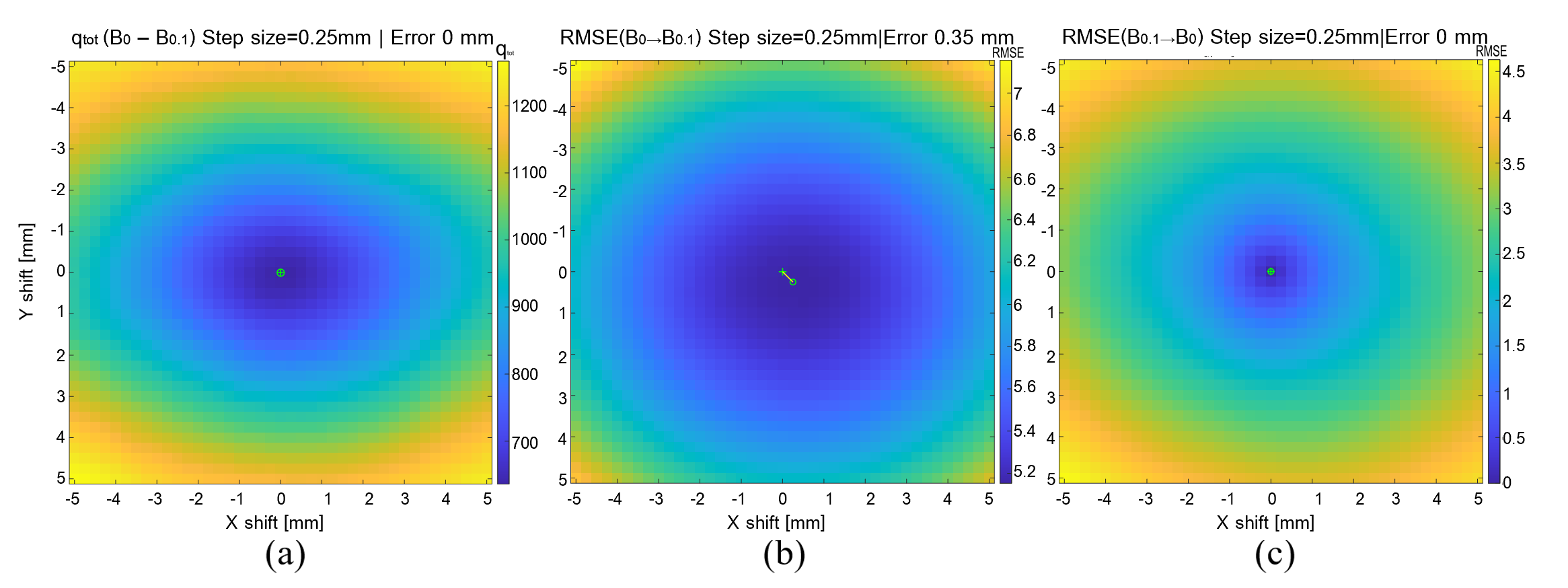} %
    \caption{Presence of differences in densities: $q_{tot}$ for displacements on the XY plane in ROI [-5 +5] mm (a), and RMSE for the case $B_0$ → $B{0.1}$ (b) and $B{0.1}$ → $B_0$ (c).}
    \label{F7}
\end{figure}

Adopting $a=1$, the trend of $q_{tot}$ as a function of point cloud density is presented. Specifically, additional point clouds were created by performing random downsampling of $B_0$ at 90\%, 70\%, 50\%, 30\%, and 10\%, named respectively $B_{0.9}$, $B_{0.7}$, $B_{0.5}$, $B_{0.3}$, and $B_{0.1}$. Fig.~\ref{F7} presents the results as colormaps of $q_{tot}$ (Fig.~\ref{F7}a), and of RMSE for the case $B_0 \rightarrow B_{0.1}$ (Fig.~\ref{F7}b) and $B_{0.1} \rightarrow B_0$ (Fig.~\ref{F7}c). The $B_0$ – $B_{0.1}$ comparison was chosen as it represents the most challenging scenario.

In the $q_{tot}$ colormap (Fig.~\ref{F7}a), the center of the ring of maximum values is shown. The central point, identified by the green cross (representing perfect alignment), overlaps with the pixel representing the translation with the minimum $q_{tot}$ value (green circle). The distance between these two points, which is the calculated error relative to the ground truth, is therefore zero. The error is defined as the distance from the actual perfect alignment position to the position the point cloud would assume if registered using the selected metric ($q_{tot}$ or RMSE). A zero error indicates that the alignment achieved using one of the metrics would coincide with a perfect alignment.

The same observation applies to the RMSE colormap for $B_{0.1} \rightarrow B_{0}$. In fact, it is a good practice to compute distances from the less dense point cloud to the denser one. Otherwise, Fig.~\ref{F7}b (RMSE $B_{0} \rightarrow B_{0.1}$) shows a shift between the ground truth (green cross) and the minimum RMSE value (green circle) of 0.36 mm. This supports the fact that $q_{tot}$ has a minimum exactly at the perfect alignment, and it highlights the importance of choosing the moving and fixed point clouds in algorithms like those based on RMSE. A key strength of the method proposed in this paper is that, although in this testing phase the less dense point cloud is known, in a real case scenario, a point cloud might have variable local density.

\subsection{Influence of noise on $q_{tot}$}
Starting from $B_{0}$, additional point clouds were generated by introducing random noise within the volume of the bounding parallelepiped that ideally contains the entire point cloud, with progressively increasing percentages. Specifically, the point clouds $B_{n0.1}$, $B_{n0.2}$, $B_{n0.3}$, and so on were generated. The distance $r_{4th}$ was evaluated similarly to the previous case. 
\begin{figure}[h] 
    \centering
    \includegraphics[width=\columnwidth]{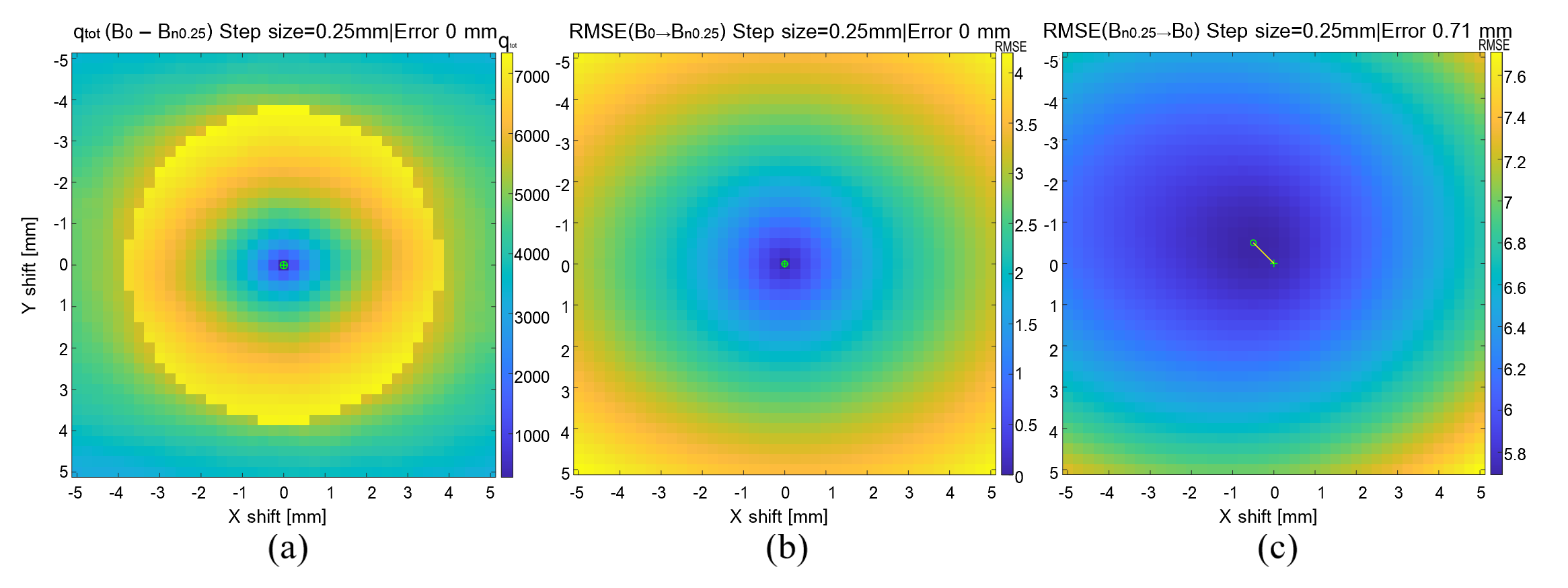} %
    \caption{Presence of noise: $q_{tot}$ for displacements on the XY plane in ROI [-5 +5] mm (a), and RMSE for the case $B_0$ → $B_n{0.25}$ (b) and $B_n{0.25}$ → $B_0$ (c).}
    \label{F8}
\end{figure}
Similar considerations can be made as in the previous case. $q_{tot}$ and RMSE ($B_{0} \rightarrow B_{n0.25}$), represented in Fig. 8a and Fig. 8b respectively, also show a minimum in the case of perfect alignment in the presence of noise, while the RMSE of $B_{n0.25} \rightarrow B_{0}$ (Fig. 8c) indicates an error of 0.71 mm. 
\subsection{Influence of Holes on $q_{tot}$}
\begin{figure}[h] 
    \centering
    \includegraphics[width=\columnwidth]{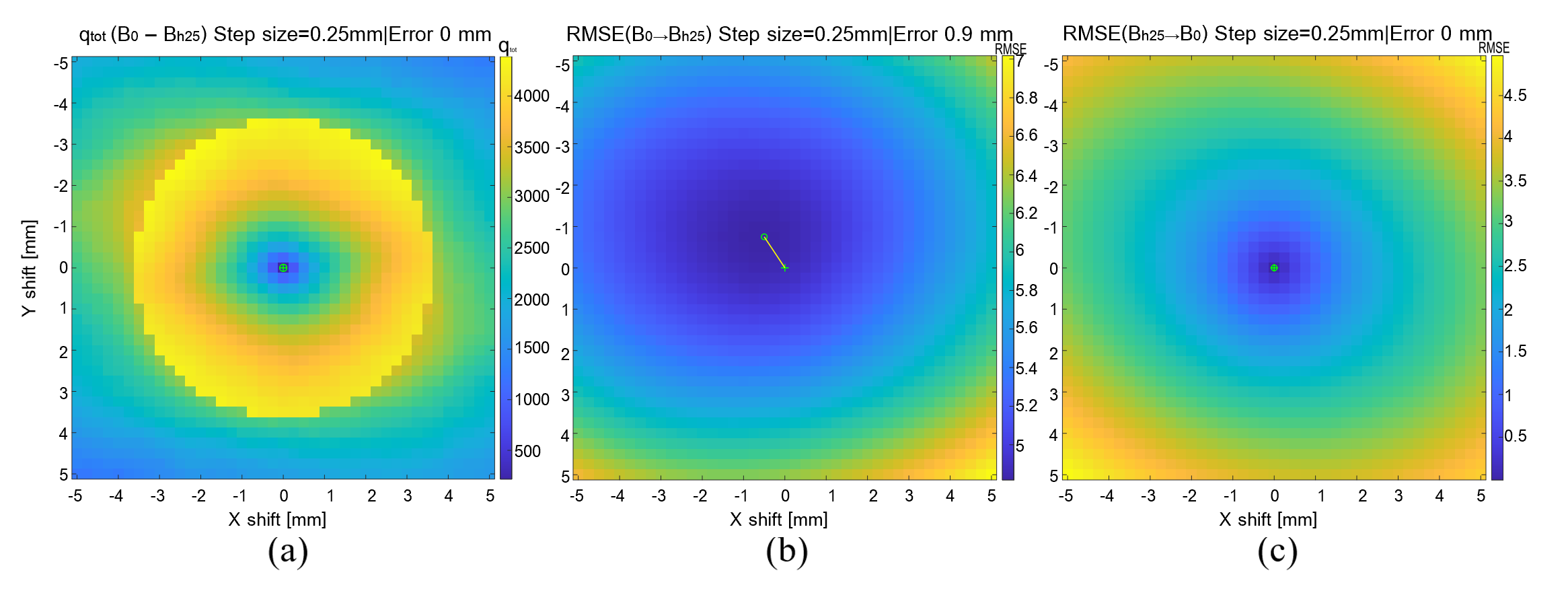} %
    \caption{Presence of holes: $q_{tot}$ for displacements on the XY plane in ROI [-5 +5] mm (a), and RMSE for the case $B_0$ → $B_h{25}$ (b) and $B_h{25}$ → $B_0$ (c).}
    \label{F9}
\end{figure}

Following the same approach as before, the behavior of $q_{tot}$ was evaluated in relation to the presence of holes in the point cloud. Consequently, starting from $B_{0}$, five point clouds with holes were generated. The holes were created by removing sections of the point cloud. Starting from a random selection of 30 points (seeds) in the point cloud, points nearby were incrementally eliminated. Specifically, neighbors were removed in numbers of 5, 10, 15, 20, and 25, resulting in the respective point clouds $B_{h5}$, $B_{h10}$, $B_{h15}$, $B_{h20}$, and $B_{h25}$. The colormaps are presented (Fig.~\ref{F9}).
Even in the presence of holes, similar considerations to those previously discussed can be made. It is observed that an error of 0.9 mm is recorded for the RMSE ($B_0 \rightarrow B_{h25}$).
\subsection{Trends of $q_tot$ for Partially Overlapping Point Clouds}
\begin{figure}[h] 
    \centering
    \includegraphics[width=0.6\columnwidth]{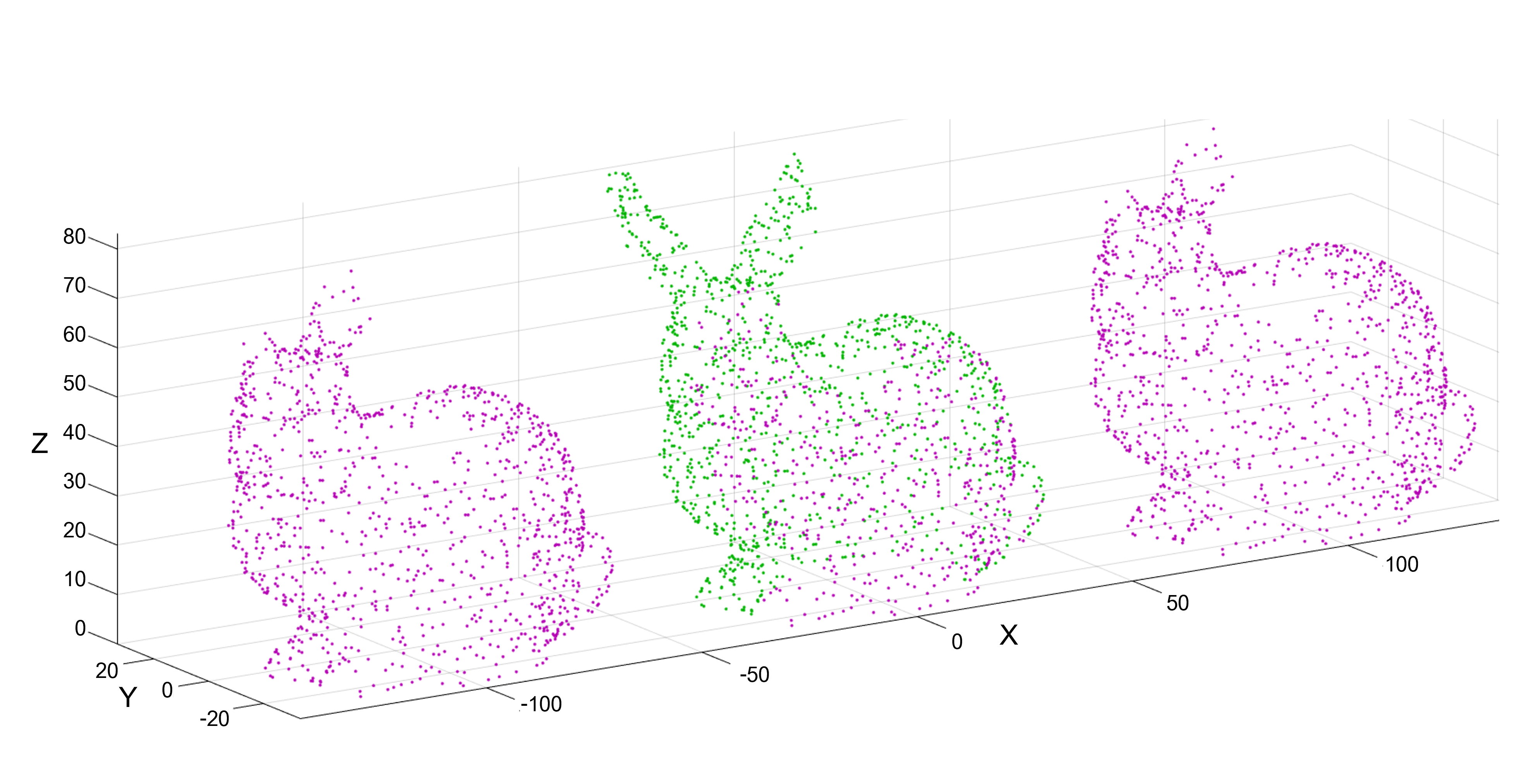} %
    \caption{$B_{0p1}$ (green) and $B_{0p2}$ (magenta) and simulation of displacements along the X-axis of $B_{0p2}$.}
    \label{F10}
\end{figure}

The behavior of $q_{tot}$ was then evaluated in the case of two partially overlapping point clouds, such as in the scenario of a partial acquisition from two views of a target. The point clouds ($B_{0p1}$ and $B_{0p2}$) were generated by taking two sections of the point cloud $B_0$ so that they were partially overlapping. Fig.~\ref{F10} displays both point clouds, with different colours highlighting their differences ($B_{0p1}$ is shown in green and $B_{0p2}$ in magenta). 

\begin{figure}[h] 
    \centering
    \includegraphics[width=\columnwidth]{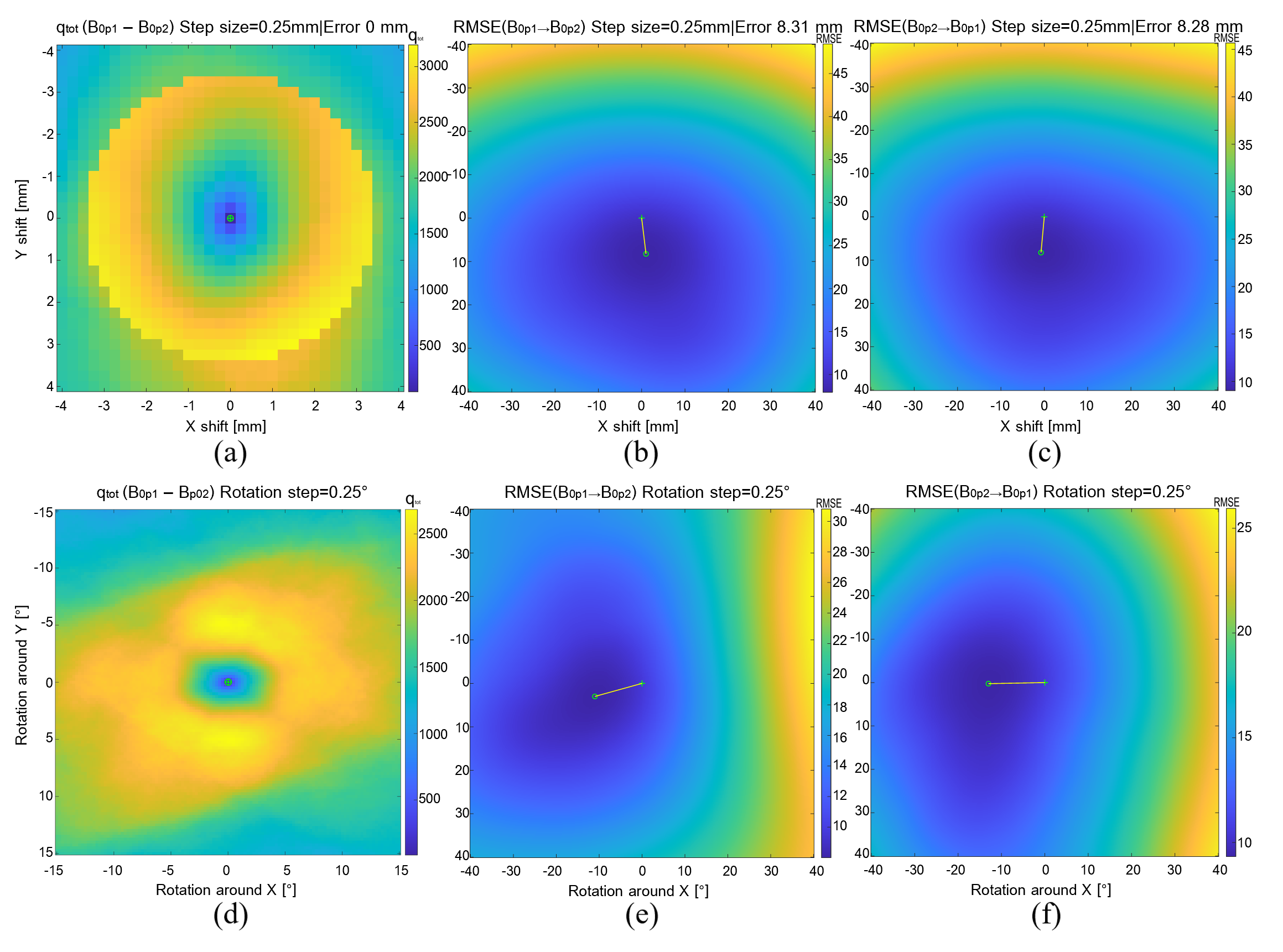} %
    \caption{Partial overlap: $q_{tot}$ for displacements on the XY plane in the ROI [-5 +5] mm (a), and RMSE for the case $B_p{01} \rightarrow B_p{02}$ (b) and $B_p{02} \rightarrow B_p{01}$ (c). $q_{tot}$ for rotations around the barycentric X and Y axes in the ROI [-5 +5] mm (a), and RMSE for the case $B_p{01} \rightarrow B_p{02}$ (b) and $B_p{02} \rightarrow B_p{01}$ (c).}
    \label{F11}
\end{figure}

Below (Fig.~\ref{F11}), the colormap for $q_{tot}$ and the RMSE for translations in the XY plane and rotations around the centroids' X and Z axes are shown.
For the translations on the XY plane, shown in Fig.~\ref{F11}a., it can be observed that $q_{tot}$ exhibits a minimum at the point of perfect alignment (central pixel). In contrast, in Fig.~\ref{F11}b. and Fig.~\ref{F11}c., the RMSE shows an error in both $B_{0p1} \rightarrow B_{0p2}$ and $B_{0p2} \rightarrow B_{0p1}$ cases. The measured errors in these latter cases are 8.31 mm and 8.28 mm, respectively, which are significantly higher than all previously explored cases. 

Finally, the colormaps in Fig.~\ref{F11}d, Fig.~\ref{F11}e, and Fig.~\ref{F11}f display the values of $q_{tot}$, RMSE ($B_{0p1} \rightarrow B_{0p2}$), and RMSE ($B_{0p2} \rightarrow B_{0p1}$) respectively, representing the rotations around the centroids' X and Y axes with a rotation step of 0.25$^{\circ}$. In the case of $q_{tot}$, the perfect alignment corresponds to the minimum value of $q_{tot}$, surrounded by a ring of maxima. The RMSE, in both cases, has an absolute minimum value that should correspond to perfect alignment, which is distant from the actual alignment.
\subsection{Behavior of $q_{tot}$ for Similar Point Clouds}
\begin{figure}[h] 
    \centering
    \includegraphics[width=0.5\columnwidth]{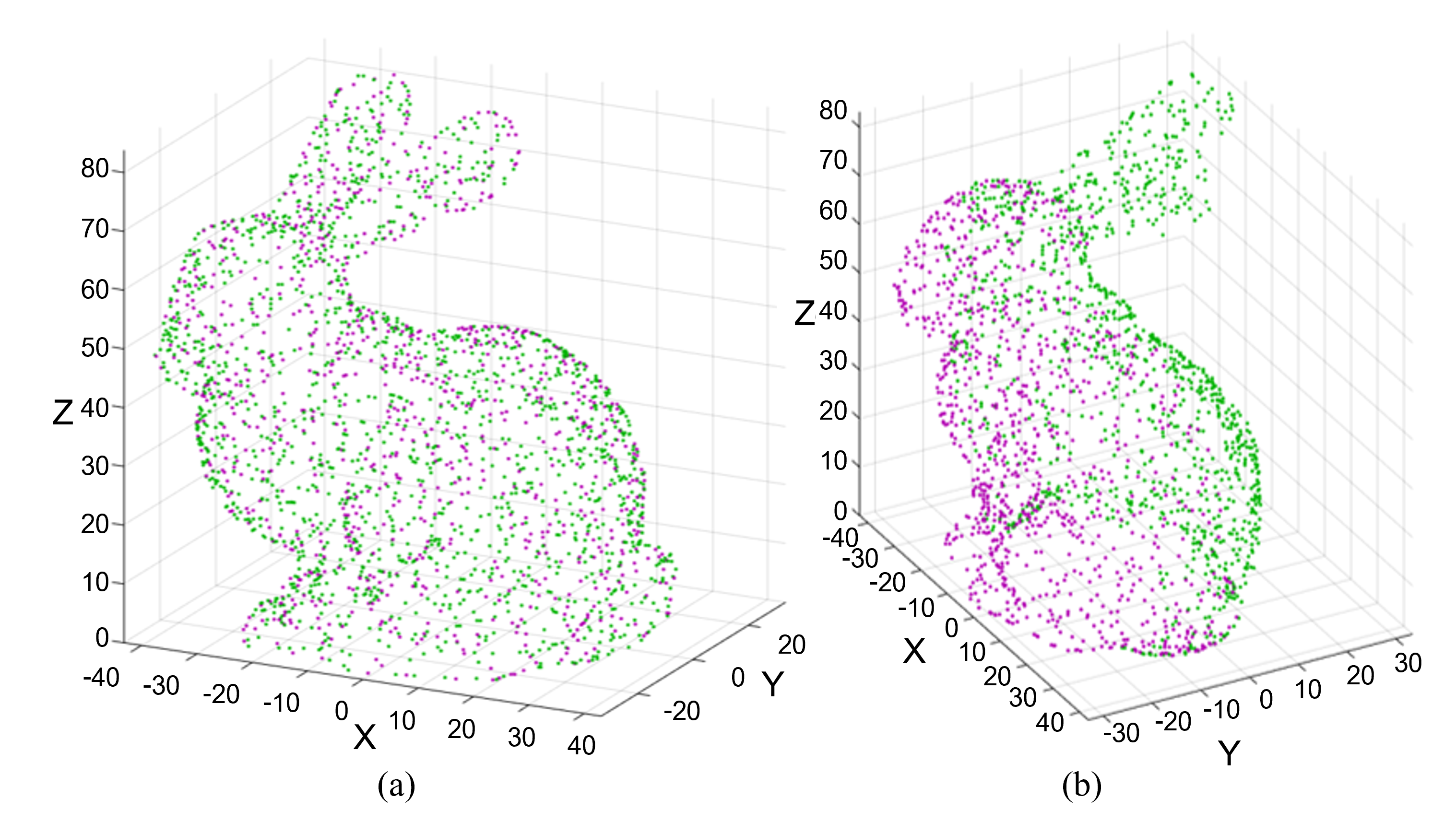} %
    \caption{(a) $B_{0}$ (green) and $B_{0r}$ (magenta). (b) $B_{0p1}$ (green) and $B_{0rp}$ (magenta)}
    \label{F12}
\end{figure}

The values of $q_{tot}$ and RMSE were subsequently evaluated for the alignment of two point clouds that, while representing the same object, differ in mesh. Therefore, a remesh (Isotropic Explicit 
Remeshing \cite{58}) was performed on $B_0$ using the MeshLab software (v. 2023.12) \cite{59}. The resulting point cloud will 
be referred to as $B_{0r}$, represented in green in Fig.~\ref{F12}(a).

\begin{figure}[h] 
    \centering
    \includegraphics[width=\columnwidth]{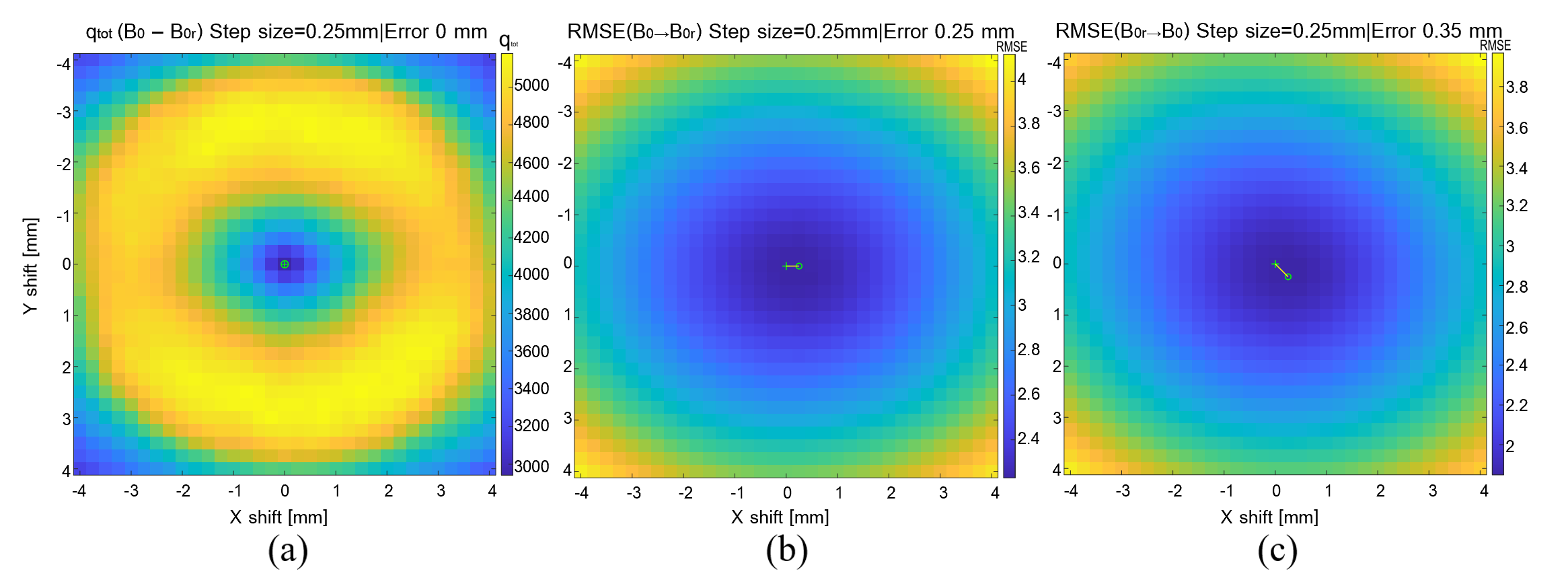} %
    \caption{Similar point clouds: $q_{tot}$ for displacements on the XY plane in the ROI [-5 +5] mm (a), and RMSE for the case $B_{0} \rightarrow B_{0r}$ (b) and $B_{0r} \rightarrow B_{0}$ (c).}
    \label{F13}
\end{figure}

The results of the XY plane translations are now presented using colormaps for the comparison of $q_{tot}$ and RMSE (Fig.~\ref{F13}).
The colormaps show results similar to those already presented. The RMSE error is present in both studied cases, $B_0 \rightarrow B_{0r}$ and $B_{0r} \rightarrow B_0$, with values of 0.25 mm and 0.35 mm, respectively.
\subsection{Behavior of $q_{tot}$ for Similar Point Clouds with Partial Overlap}
\begin{figure}[h] 
    \centering
    \includegraphics[width=\columnwidth]{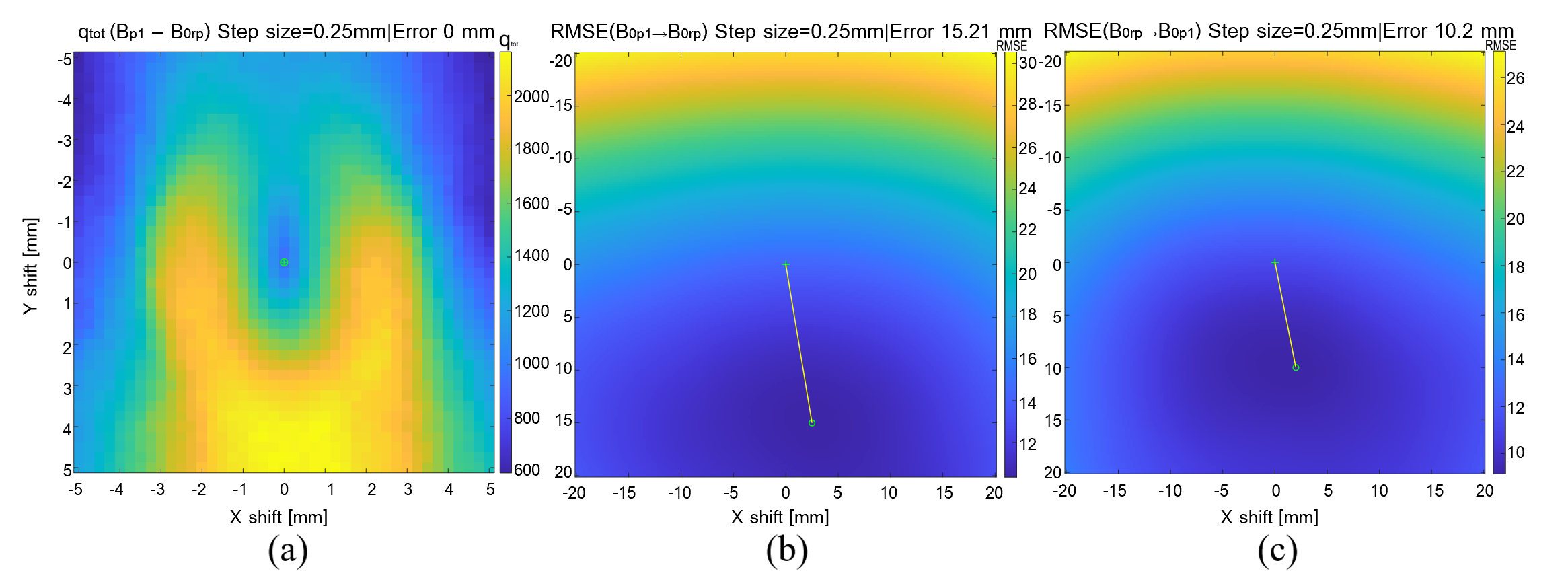} %
    \caption{Partial overlap of two similar point clouds: $q_{tot}$ for displacements on the XY plane in the ROI [-5 +5] mm (a), and RMSE for the case $B_{0p1} \rightarrow B_{0rp}$ (b) and $B_{0rp} \rightarrow B_{0p1}$ (c).}
    \label{F15}
\end{figure}

Finally, a more complex case is investigated, involving the presence of two similar point clouds that are partially overlapping. The point clouds used are $B_{0p1}$, while the complementary half has been extracted from $B_{0r}$ and will be referred to as $B_{0rp}$. Fig.~\ref{F12}(b) shows the two point clouds ($B_{0p1}$ and $B_{0rp}$) and Fig.~\ref{F15} displays the results using colormaps.

\begin{figure}[h!] 
    \centering
    \includegraphics[width=0.66\columnwidth]{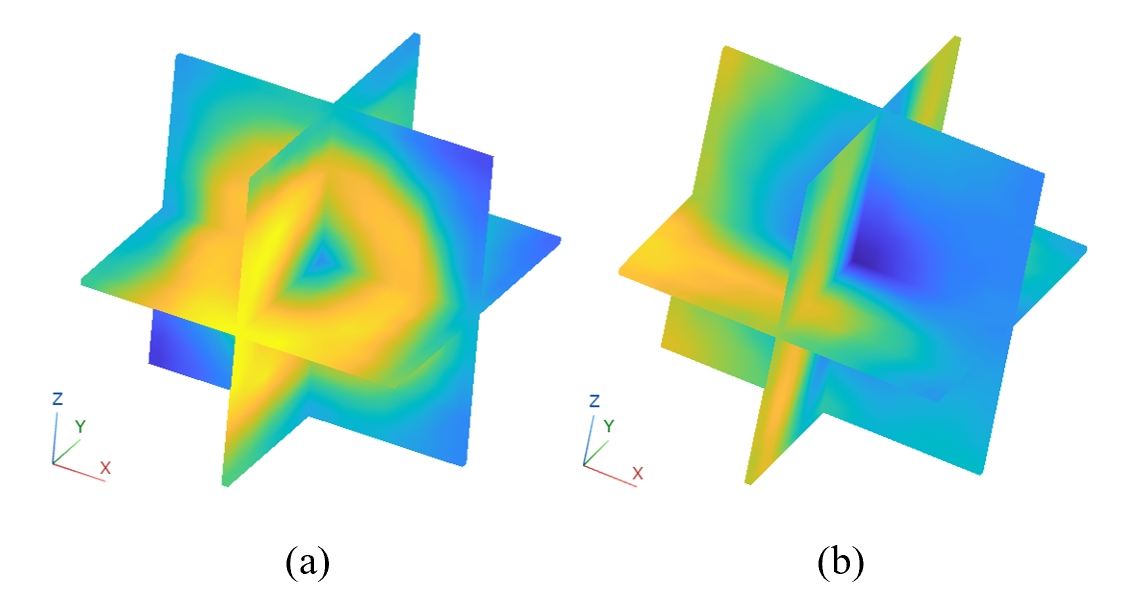} %
    \caption{$q_{tot}$ for translations within the ROI on the XY, XZ, and YZ planes (a) and for rotation around centroids' X, Y, and Z axes (b).}
    \label{F16}
\end{figure}

From the $q_{tot}$ colormap (Fig.~\ref{F15}a), it can be observed that the amplitude of the maxima around the central pixel, which represents the point of perfect alignment, has decreased, losing some of its crown shape. However, although less pronounced, maxima are still present around the central pixel, which in this case assumes the minimum value. The RMSE colormaps (Fig.~\ref{F15}b and Fig.~\ref{F15}c) show a significant deviation between the detected minimum point and the ground truth. In fact, in the case of $B_{0p1} \rightarrow B_{rp}$, the error is 15.21 mm, while in the case of $B_{rp} \rightarrow B_0$, it is 10.20 mm, the highest recorded so far. For completeness, it is also shown how $q_{tot}$ changes in response to translations and relative rotations between the two point clouds, specifically within the XYZ space (Fig.~\ref{F16}a) and around the X, Y, and Z axes (Fig.~\ref{F16}b).

\begin{figure}[h!] 
    \centering
    \includegraphics[width=0.75\columnwidth]{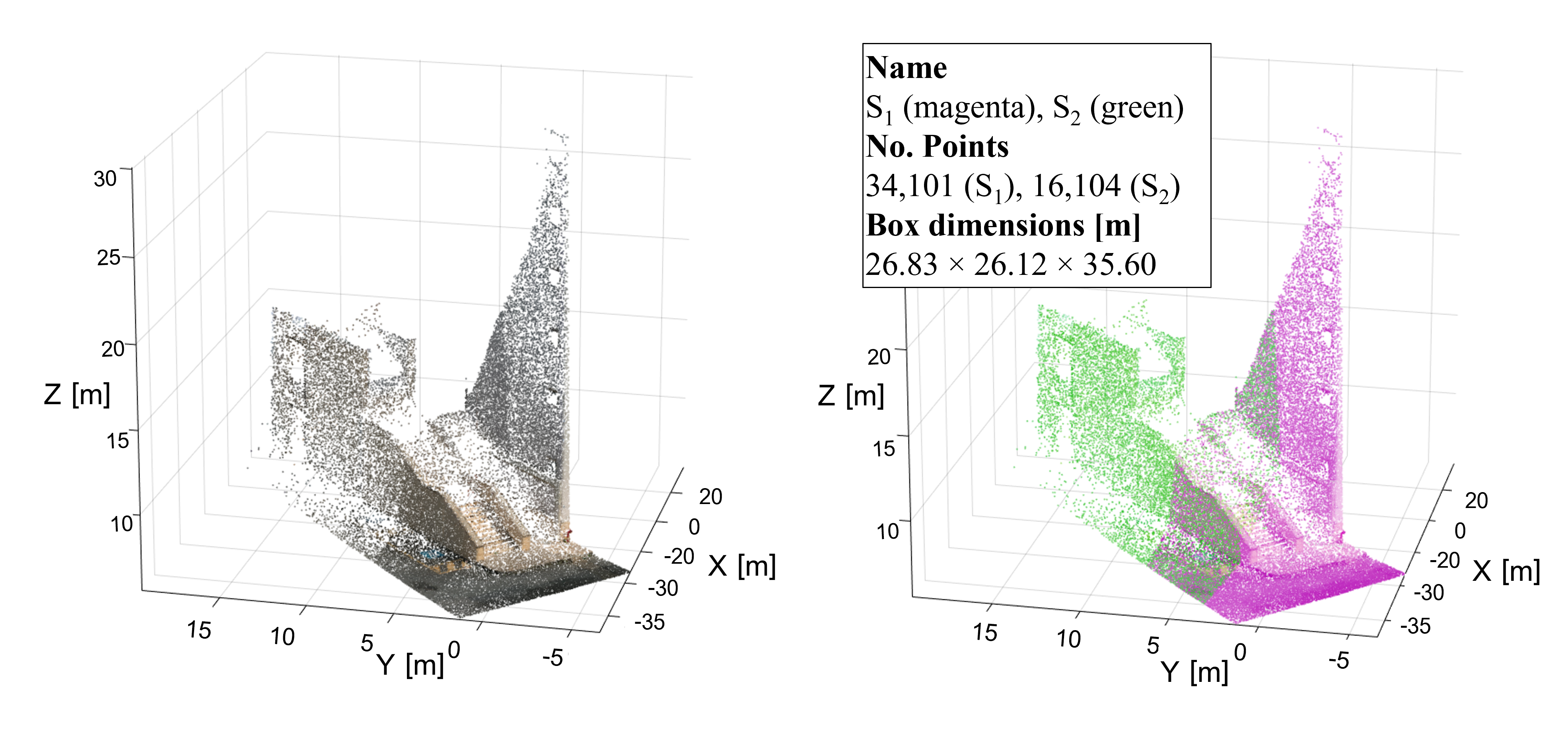} %
    \caption{LiDAR point cloud: colored representation and distinction between $S_1$ (magenta) and $S_2$ (green)}
    \label{F17}
\end{figure}

For a more comprehensive analysis, the behavior of a different type of point cloud was investigated, both in terms of representation and the number of points. The chosen point cloud represents an outdoor scene captured using LiDAR (external masonry staircase of a building, captured in an urban context). The most challenging and realistic case was investigated, which involves two partially overlapping point clouds. These were obtained through two random downsampling processes of an original point cloud and then cropped to create a partial overlap. The two point clouds are defined as $S_1$ and $S_2$ (Fig.~\ref{F17}), with 34,101 and 16,104 points, respectively. Therefore, the two point clouds have different numbers of points, different local densities, and also contain fragmented data due to imperfect LiDAR acquisition. The value of $q_{tot}$ (and RMSE) was then calculated as a function of translations on the XY.
\begin{figure}[!h] 
    \centering
    \includegraphics[width=\columnwidth]{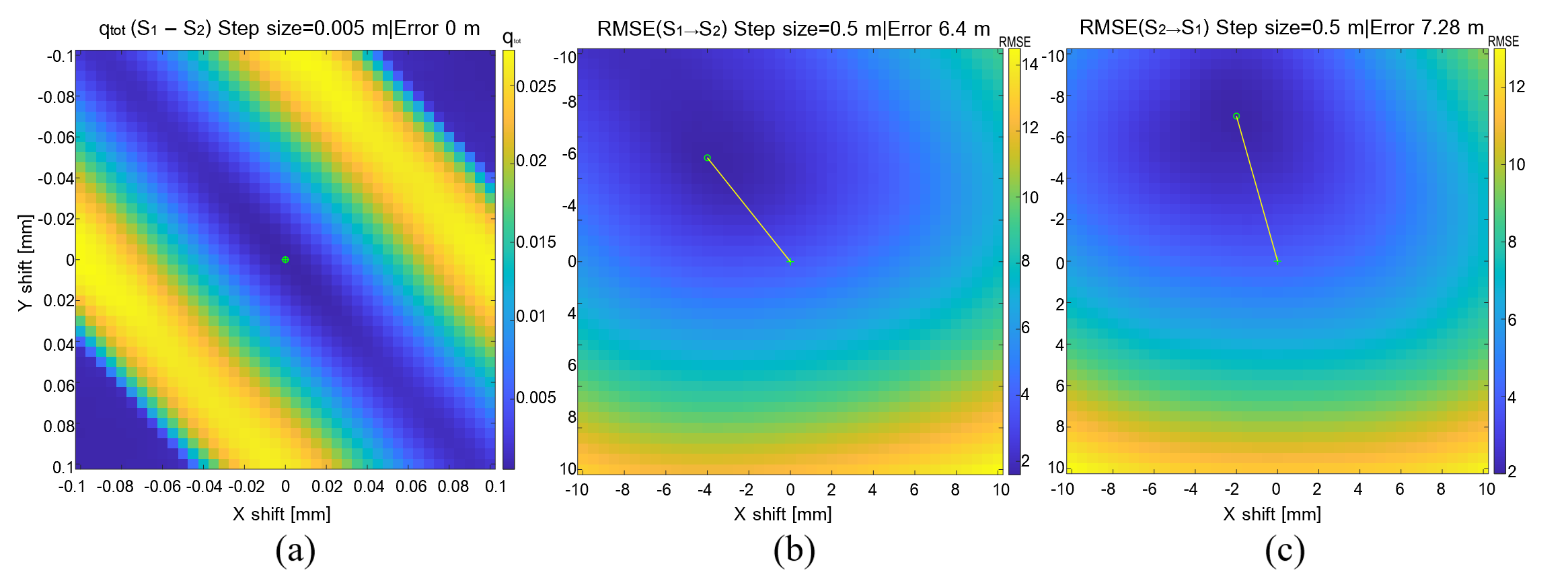} %
    \caption{Scene point clouds partially overlapping: $q_{tot}$ for displacements on the XY plane in the ROI [-0.1 +0.1] m (a), and RMSE for the case $S_{1} \rightarrow S_{2}$ (b) and $S_{2} \rightarrow S_{1}$ (c) in [-10 +10] m.}
    \label{F18}
\end{figure}
\begin{table*}[!htb]
    \centering
    \caption{Summary of the Experiments.}
    \setlength{\tabcolsep}{3pt}  
    \renewcommand{\arraystretch}{1.2}  
    \begin{tabular}{|p{1.6cm}|p{4.8cm}|p{2cm}|p{1cm}|p{1.2cm}|p{2.5cm}|p{1.4cm}|p{1.5cm}|}
        \hline
        \textbf{Comparison} & \textbf{Description} & \textbf{No. points} & \textbf{$r_{4th}$ [mm]} & \textbf{$q_{tot}$ error [mm]} & \textbf{RMSE error [mm] $P_1 \rightarrow P_2$, $P_2 \rightarrow P_1$} & \textbf{Chamfer error [mm]} & \textbf{Hausdorff error [mm]}\\
        \hline
        $B_0 - B_0$ & $B_0$ with $B_0$ & 1,597 -- 1,597 & 3.13 & 0 & 0, 0 & 0 & 0 \\
        $B_0 - B_{0.9}$ & $B_0$ with $B_0$ sampled at 90\% & 1,597 -- 1,437 & 3.50 & 0 & 0, 0 & 0 & 1.6 \\
        $B_0 - B_{0.7}$ & $B_0$ with $B_0$ sampled at 70\% & 1,597 -- 1,118 & 3.79 & 0 & 0, 0 & 0 & 0.79 \\
        $B_0 - B_{0.5}$ & $B_0$ with $B_0$ sampled at 50\% & 1,597 -- 799 & 4.30 & 0 & 0, 0 & 0 & 1.52 \\
        $B_0 - B_{0.3}$ & $B_0$ with $B_0$ sampled at 30\% & 1,597 -- 479 & 5.50 & 0 & 0.25, 0 & 0 & 2.46\\
        $B_0 - B_{0.1}$ & $B_0$ with $B_0$ sampled at 10\% & 1,597 -- 160 & 9.67 & 0 & 0.35, 0 & 0 & 3.04\\
        $B_0 - B_{n0.05}$ & $B_0$ with $B_0$ + 5\% noise & 1,597 -- 1,677 & 3.55 & 0 & 0, 0 & 0 & 10.82\\
        $B_0 - B_{n0.10}$ & $B_0$ with $B_0$ + 10\% noise & 1,597 -- 1,757 & 3.68 & 0 & 0, 0.25 & 0 & 13.44\\
        $B_0 - B_{n0.15}$ & $B_0$ with $B_0$ + 15\% noise & 1,597 -- 1,837 & 3.73 & 0 & 0, 0.56 & 0 & 7.39\\
        $B_0 - B_{n0.20}$ & $B_0$ with $B_0$ + 20\% noise & 1,597 -- 1,916 & 3.77 & 0 & 0, 0.36 & 0 & 14.98\\
        $B_0 - B_{n0.25}$ & $B_0$ with $B_0$ + 25\% noise & 1,597 -- 1,996 & 3.80 & 0 & 0, 0.71 & 0 & 8.28\\
        $B_0 - B_{h5}$ & $B_0$ with $B_0$ + holes (5 points per seed) & 1,597 -- 1,447 & 3.45 & 0 & 0 ,0 & 0 & 3.55\\
        $B_0 - B_{h10}$ & $B_0$ with $B_0$ + holes (10 points per seed) & 1,597 -- 1,297 & 3.50 & 0 & 0, 0 & 0 & 6.10\\
        $B_0 - B_{h15}$ & $B_0$ with $B_0$ + holes (15 points per seed) & 1,597 -- 1,147 & 3.51 & 0 & 0, 0 & 0 & 4.56\\
        $B_0 - B_{h20}$ & $B_0$ with $B_0$ + holes (20 points per seed) & 1,597 -- 997 & 3.50 & 0 & 0.35, 0 & 0 & 7.27\\
        $B_0 - B_{h25}$ & $B_0$ with $B_0$ + holes (25 points per seed) & 1,597 -- 847 & 3.62 & 0 & 0.90, 0 & 0 & 5.27\\
        $B_{0p1} - B_{0p2}$ & $B_0$ partial 1 with $B_0$ partial 2 & 1,087 -- 1,112 & 3.44 & 0 & 8.31, 8.28 & 0 & 23.69\\
        $B_0 - B_{0r}$ & $B_0$ with remeshed $B_0$ & 1,597 -- 968 & 4.04 & 0 & 0.25, 0.35 & 0 & 2.46\\
        $B_{0p1} - B_{rp}$ & $B_0$ partial 1 with remeshed $B_0$ partial & 1,087 -- 967 & 3.42 & 0 & 15.21, 10.20 & 11.92 & 24.14\\
        $S_{1} - S_{2}$ & S partial 1 with S partial 2 & 34,101 -- 16,101 & 0.14 [m] & 0 & 6.40, 7.28 [m] & 6.02 [m] & 5.32 [m]\\
        \hline
    \end{tabular}
    \label{T2}
\end{table*}
\subsection{Sensitivity Analysis}
\begin{table}[!h]
    \centering
    \caption{Sensitivity Analysis.}
    \setlength{\tabcolsep}{3pt}  
    \renewcommand{\arraystretch}{1.2}  
    \begin{tabular}{|p{1cm}|p{1cm}|p{1cm}|p{1cm}|}
        \hline
        \textbf{$\sigma_{noise}$} & \textbf{$\mu_{q_{tot}}$} & \textbf{$\sigma_{q_{tot}}$} & \textbf{$CV$} \\
        \hline
        0.01 & 48.83   & 5.89   & 0.12\\
        0.02 & 138.39  & 8.80   & 0.06 \\
        0.05 & 464.30  & 15.90  & 0.03 \\
        0.1  & 972.78  & 26.03  & 0.03 \\
        \hline
    \end{tabular}
    \label{T1}
\end{table}

\begin{figure}[!h] 
    \centering
    \includegraphics[width=0.6\columnwidth]{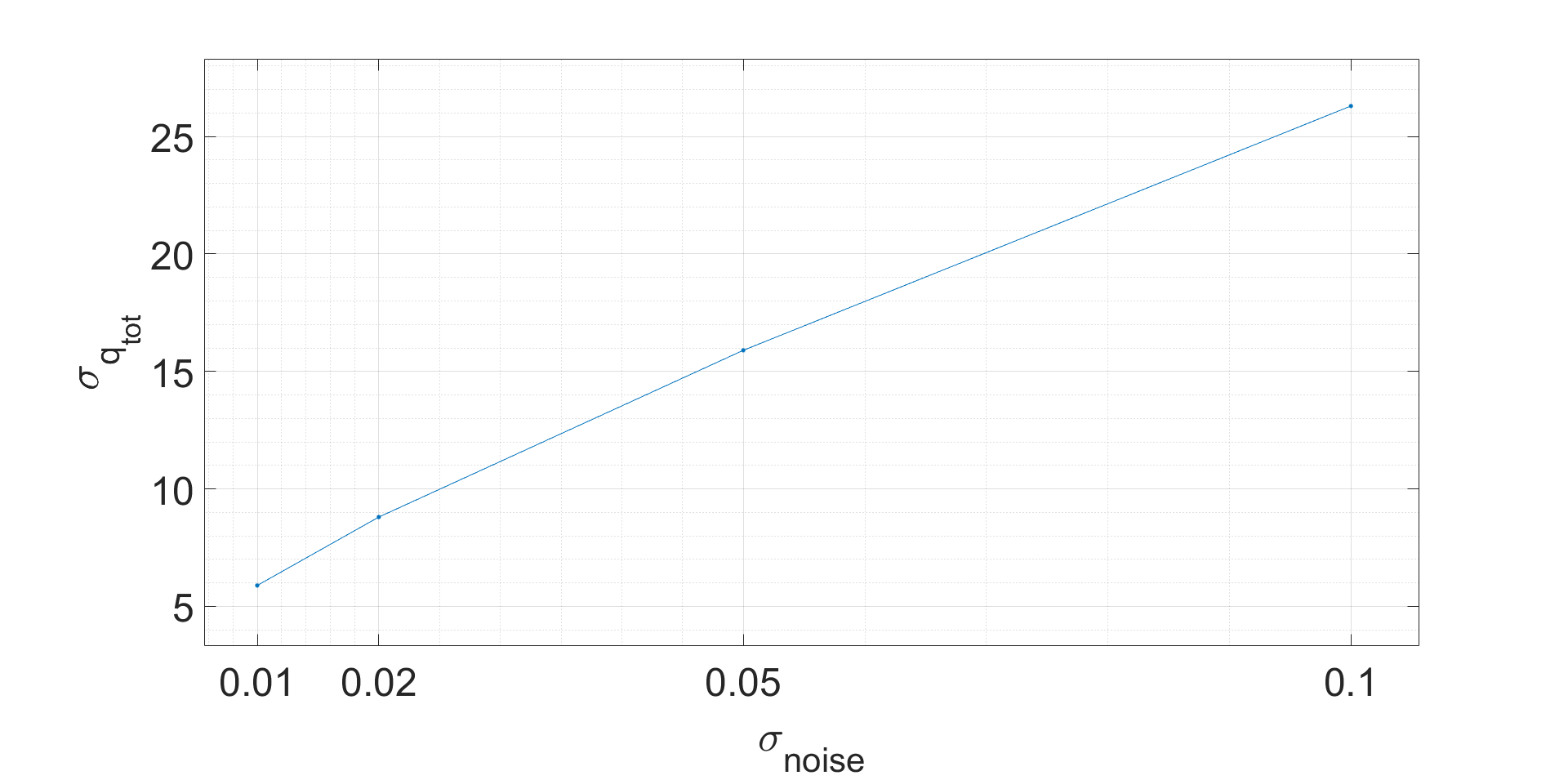} %
    \caption{Graph of $\sigma_{q_{tot}}$ as a function of $\sigma_{noise}$}
    \label{F19}
\end{figure}

A sensitivity analysis of the method was performed using the Monte Carlo approach. Specifically, taking $B_0$ as a reference, 1,000 point clouds were generated. Each generated point cloud was subjected to random Gaussian noise. Each point was randomly displaced from its initial position following a Gaussian distribution with different standard deviations ($\sigma_{noise}$), to simulate the potential errors introduced by the acquisition instrument. With this analysis, it is possible to assess how these data variations affect the value of $q_{tot}$. Table~\ref{T1} presents the mean values ($\mu_{q_{tot}}$), standard deviation ($\sigma_{q_{tot}}$), and coefficients of variation ($CV$) obtained using the Monte Carlo method. Fig.~\ref{F19} shows the behavior of $\sigma_{q_{tot}}$ as $\sigma_{noise}$ increases and illustrates how, within the error range, the growth of $\sigma_{q_{tot}}$ remains stable. The sensitivity analysis reveals that the mean values of $q_{tot}$ increase as the perturbation increases, consistently with its definition. The $CV$ is very low and tends to decrease as the perturbation increases, maintaining sufficiently low values. This relatively low variability indicates a sufficient robustness of the proposed metric to noise.

\subsection{Summary of the Experiments}
Table~\ref{T2} provides a summary of the experiments conducted and the results obtained for translations on the XY plane. It can be observed that minimizing $q_{tot}$ always results in perfect alignment, while minimizing the RMSE may yield a result different from perfect alignment. This was observed, for example, when the density difference between the two point clouds becomes large (see the $B_0 \rightarrow B_{0.1}$ case), or in the presence of both noise and holes. Furthermore, since RMSE calculation is not commutative, the choice of which point cloud to transform affects the outcome. The highest errors occur in the last three cases listed in Table~\ref{T1}: identical and partially overlapping point clouds, similar point clouds, and similar, partially overlapping point clouds (which is one of the most practically relevant cases). In these cases, both RMSE errors are non-zero, regardless of the choice of which point cloud to transform.
Finally, for further comparison of the results, the proposed metric is also evaluated against the Chamfer distance and the Hausdorff distance (Table~\ref{T2}). It can be observed that the Chamfer distance performs well in the various comparisons, except in the case of partial overlap between two non-identical point clouds. As for the Hausdorff distance, the error is zero only when the two point clouds are identical and perfectly overlapped.

\section{Conclusions}\label{sec:4}
The paper provides insight into the proposed metric upon which IDEM, a method for fine rigid pairwise 3D point cloud registration, relies. The metric introduced in this paper is based on a novel formulation of the differential entropy of point clouds, which was previously introduced by the research group in earlier works. Experimental results demonstrate that this metric, $q_{tot}$, can address several limitations associated with point cloud registration methods that rely on Euclidean distance-based approaches.
The efficiency of $q_{tot}$ has been shown in identifying optimal alignment under varying conditions, including point clouds with different densities, noise, holes, similar shapes, and cases of partial overlap. Comparisons were made using colormaps (both 2D and 3D) simulating translations and rotations along the three axes, in conjunction with RMSE, a commonly used metric for evaluating point cloud alignment. The function $q_{tot}$ was observed to exhibit absolute maxima around the point of perfect alignment (ROI), where the function reaches a minimum, denoted $q_{IDEM}$. Notably, these maxima are found in a neighborhood around the aligned position, corresponding to approximate values of $\pm 3$ mm to $\pm 7$ mm for translation and $\pm 10^\circ$ to $\pm 15^\circ$ for rotation.
The experiments show that RMSE behaves differently depending on which point cloud is selected for transformation, while IDEM is independent of this choice, thus addressing the issue of non-commutativity associated with Euclidean distance in point cloud comparisons, which can otherwise affect the performance of distance-based methods. Consequently, it is suggested that this differential entropy-based metric could serve as the foundation for an iterative fine rigid registration algorithm.
In practical terms, with an initial pre-alignment placing the point clouds within the ROI, the IDEM method can effectively leverage the neighborhood defined by the maxima of the $q_{tot}$ function. Alternatively, once the maxima are identified, a targeted algorithm could restrict the search for the function's minimum to the ROI. In both approaches, iteratively minimizing the $q_{tot}$ function would converge to $q_{IDEM}$, achieving the optimal alignment. While the present study focuses on the analytical behavior of the proposed differential entropy–based metric, future research may investigate its integration as a loss or feature within learning-based registration methods, potentially combining the interpretability of the proposed metric with the flexibility of deep neural frameworks.

\bibliographystyle{IEEEtran}      
\bibliography{references}       
\section{Biography Section}

\begin{IEEEbiography}[{\includegraphics[width=1in,height=1.25in,clip,keepaspectratio]{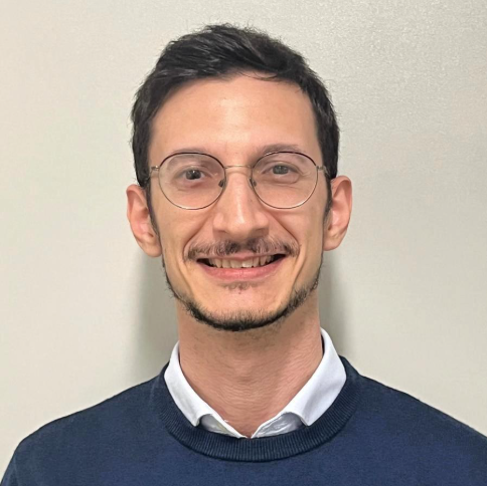}}]{Emmanuele Barberi}
is a Research Fellow in Design Methods of Industrial Engineering at the University of Messina. He obtained his Master's degree in Mechanical Engineering in 2020 from the same institution and completed his PhD in Engineering and Chemistry of Materials and Constructions in 2023, with a thesis focused on new applications of analysis based on the differential entropy of point clouds. His research primarily focuses on the analysis of 3D geometries, particularly differential entropy analysis of 3D point clouds and point cloud registration, as well as imaging analysis, processing, and machine learning applications. The National Association of Design and Methods of Industrial Engineering (ADM) awarded first place to his Doctoral Thesis in the "3D Engineering" category. 
\end{IEEEbiography}
\begin{IEEEbiography}[{\includegraphics[width=1in,height=1.25in,clip,keepaspectratio]{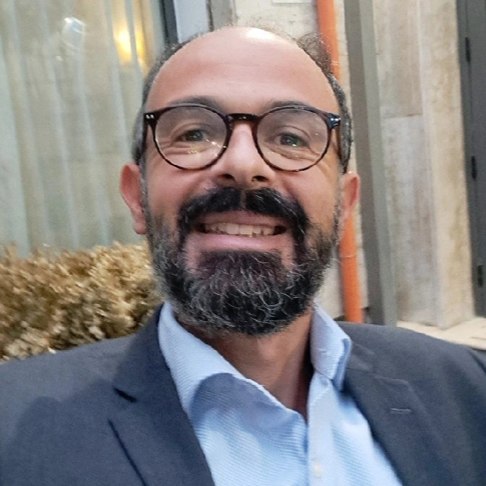}}]{Felice Sfravara}
is an Associate Professor of Design Methods of Industrial Engineering at the University of Messina. He earned a Master's degree in Naval Engineering from the University of Genoa (Italy) in 2012 and obtained his Ph.D. in Industrial Engineering from the University of Messina (Italy) in 2017, with a thesis focused on computational and experimental fluid dynamics of planing hulls. He is the author of numerous scientific papers published in international journals, covering topics such as computational life cycle assessment, environmental analyses, fluid dynamics, image analysis, virtual modeling, photogrammetry, and topology optimization. Felice Sfravara is also the inventor of the international patent application no. PCT/IT2019/000051, filed on June 18, 2018, with the University of Messina as the applicant, titled 'Gliding Hull with Motor Gas Insufflation in Water.
\end{IEEEbiography}
\begin{IEEEbiography}[{\includegraphics[width=1in,height=1.25in,clip,keepaspectratio]{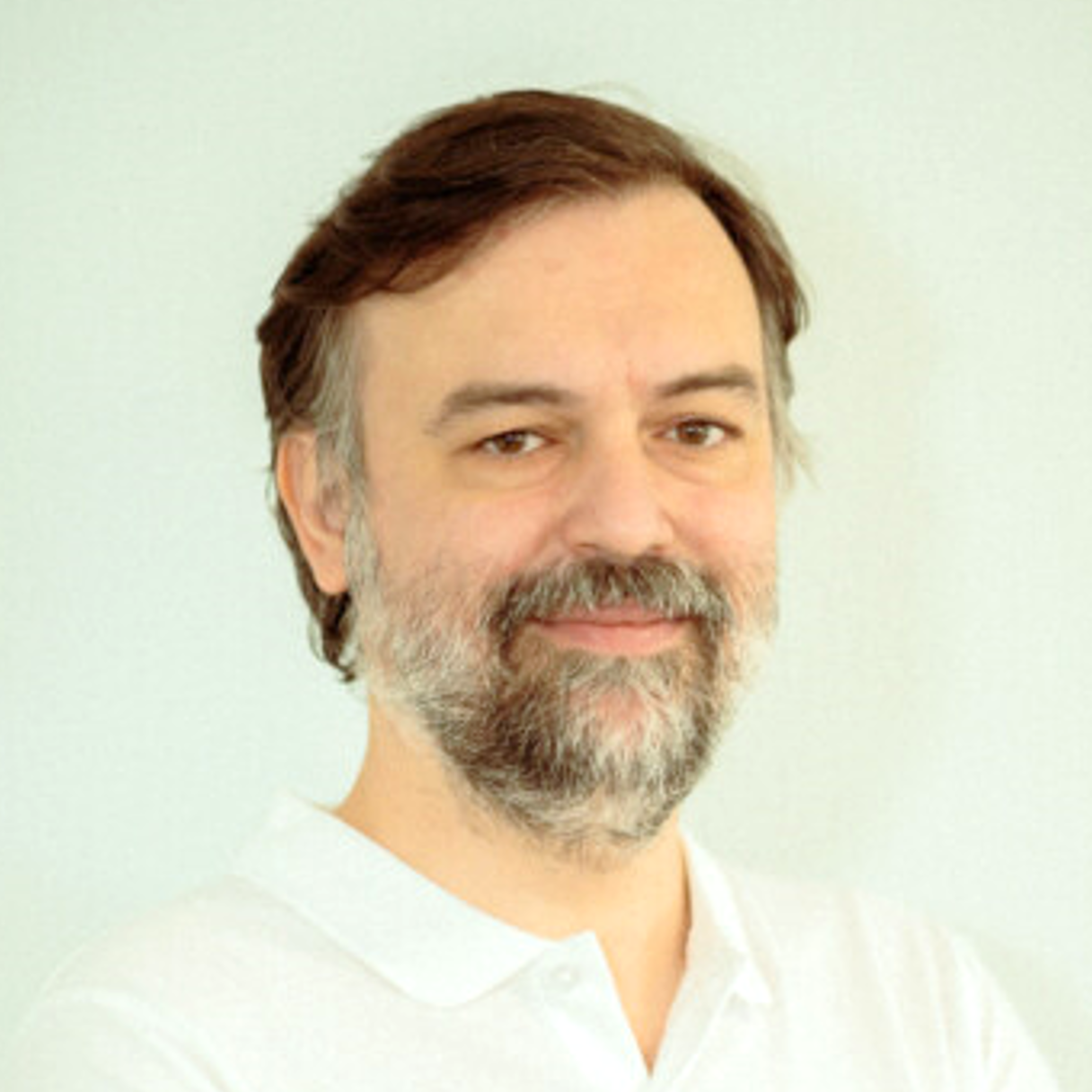}}]{Filippo Cucinotta}
is a Full Professor at the University of Messina, specialized in Design Methods of Industrial Engineering. He earned a Master's degree in Naval Engineering from the University of Genoa (Italy) in 2006 and obtained his Ph.D. in Industrial Engineering from the University of Messina (Italy) in 2009, with a thesis focused on computational and experimental fluid dynamics of planing hulls. He is the author of numerous scientific papers published in international journals, covering topics such as computational life cycle assessment, environmental analyses, fluid dynamics, image analysis, virtual modeling, photogrammetry, and topology optimization. Prof. Cucinotta is also the inventor of the international patent application no. PCT/IT2019/000051, filed on June 18, 2018, with the University of Messina as the applicant, titled 'Gliding Hull with Motor Gas Insufflation in Water.
\end{IEEEbiography}

\end{document}